\documentclass[11pt]{article}

\usepackage[preprint]{acl}

\usepackage{times}
\usepackage{latexsym}
\usepackage{booktabs}
\usepackage{array}
\usepackage{siunitx}
\usepackage{multirow}
\usepackage{rotating}
\usepackage{makecell}
\usepackage{xcolor}
\usepackage{colortbl}
\usepackage{arydshln}
\usepackage{amsmath}
\usepackage{amssymb}
\usepackage{soul}

\usepackage[T1]{fontenc}

\usepackage[utf8]{inputenc}

\usepackage{microtype}

\usepackage{inconsolata}
\usepackage{graphicx}

\usepackage{graphicx}

\usepackage[most]{tcolorbox}
\usepackage{listings}

\newtcblisting{promptbox}[1][]{
    enhanced,
    breakable,
    listing only,
    title={#1},
    colback=gray!3,
    colframe=black!55,
    coltitle=black,
    colbacktitle=gray!15,
    fonttitle=\bfseries,
    boxrule=0.5pt,
    arc=2pt,
    left=5pt,
    right=5pt,
    top=5pt,
    bottom=5pt,
    before skip=7pt,
    after skip=7pt,
    listing options={
        basicstyle=\ttfamily\scriptsize,
        columns=fullflexible,
        keepspaces=true,
        breaklines=true,
        breakatwhitespace=false,
        showstringspaces=false
    }
}

\title{Steering Geometry:\\ Validating Human Value Geometry in LLM Steering Space}

\author{
\textbf{Mohammad Mahdi Abootorabi\textsuperscript{\textdagger,$\spadesuit$}},
\textbf{Armin Saghafian\textsuperscript{\textdagger}},
\textbf{Ali Bazshoushtari\textsuperscript{\textdagger}},
\textbf{Hamid Rezaei\textsuperscript{\textdagger}},\\
\textbf{EunJeong Hwang\textsuperscript{\textdagger,$\spadesuit$}},
\textbf{Vered Shwartz\textsuperscript{\textdagger,$\spadesuit$}},
\textbf{Parvin Mousavi\textsuperscript{\textsection,$\spadesuit$}},
\textbf{Purang Abolmaesumi\textsuperscript{\textdagger}}\\[0.4em]
\textsuperscript{\textdagger}University of British Columbia \quad
\textsuperscript{$\spadesuit$}Vector Institute for AI\quad
\textsuperscript{\textsection}Queen's University
}

\usepackage{titlesec}

\titlespacing*{\paragraph}{0pt}{0.1ex plus 0.1ex minus 0.1ex}{0.5em}
\titleformat{\paragraph}[runin]{\normalfont\normalsize\bfseries}{\theparagraph}{0.2em}{}

\newboolean{isRed}
\setboolean{isRed}{False} 
\newcommand{\cam}[1]{%
    \ifthenelse{\boolean{isRed}}%
    {\textcolor{red}{#1}}
    {#1}
}

\begin{document}
\maketitle

\begin{abstract}
As large language models (LLMs) are increasingly deployed in alignment-sensitive contexts, activation steering has emerged as a lightweight, inference-time alternative to fine-tuning methods (e.g., RLHF, DPO) for behavioral control. However, existing work typically validates steering on isolated behaviors, leaving it unclear whether steering vectors encode coherent semantic structure or merely exploit behavior-specific shortcuts. 
We investigate whether the latent geometry of LLM steering vectors reflects theory-specified structure in human values and morality. Using Schwartz's Theory of Basic Human Values as our primary fine-grained framework, we introduce a 26K-sample benchmark covering 20 human values and analyze distribution-driven methods (e.g., CAA, SphericalSteer, ODESteer) and behavior-centric approaches (e.g., COLD-Steer, BiPO) across diverse model families and sizes.
We find that distribution-driven methods recover human value topologies aligned with theoretical predictions (Spearman $\rho$ up to 0.51, $p < 10^{-13}$). In contrast, behavior-centric methods achieve comparable steering performance but show little correlation with the expected value geometry. 
Geometric fidelity improves with model scale but drops after instruction tuning. Finally, better geometric alignment also leads to more human-consistent transfer across values: steering one value correctly lifts compatible values and suppresses opposing ones. Code and data are available at: \url{https://github.com/DeepRCL/Steering_Geometry}.

\end{abstract}

\section{Introduction}
Large language models (LLMs), trained on large-scale data and shaped through post-training and alignment, have become strong general-purpose assistants across diverse domains, from code generation~\cite{chen2021evaluating} to complex reasoning~\cite{wei2022chain}. However, as these models are used more often in alignment-sensitive settings where responses carry consequences, researchers face the challenge of making model behavior consistent with human expectations and values ~\cite{ouyang2022training,bai2022constitutional}.
The primary alignment framework, instruction tuning and reinforcement learning from human feedback (RLHF)~\cite{ouyang2022training,rafailov2023direct}, is highly effective at reshaping model behavior to match human preferences, but it relies heavily on resource-intensive annotations and struggles to scale across the diversity of global cultural and contextual values~\cite{askell2021general}. These shortcomings have pushed researchers toward mechanistic interpretability, specifically, understanding how models internally represent value-related concepts, and using that knowledge to make slight, targeted adjustments that steer and control model behavior.

Recent work in interpretability suggests that LLMs encode semantic concepts and behavioral traits as approximately linear directions in their activation space~\cite{zou2023representation,turner2023activation}. This linear representation hypothesis forms the basis of \textit{activation steering}: a family of inference-time methods that manipulate internal representations to control model behavior without additional training~\cite{li2023inference,rimsky2024steering}. Approaches range from contrastive activation addition~\cite{rimsky2024steering} and sparse decompositions~\cite{bayat2025steering}, to continuous ODE dynamics~\cite{zhao2026odesteer}, geometry-aware spherical rotations \citep{you2026spherical}, \cam{query-adaptive Schrödinger-bridge steering \citep{dalili2026conditional}}, and learned optimization~\cite{dunefsky2025one}. These techniques have recently been extended from simple binary traits such as truthfulness or harmlessness~\cite{wang2025adaptive} to richer constructs like persona and character~\cite{chen2025persona}. In practice, this means models can be steered toward concrete behavioral profiles, such as becoming more helpful, empathetic, or toxic.
Recent works focus on evaluating the impact of steering on a single value or behavior, without considering how the steering process affects the model's behavior across other values.
In contrast, in human psychology, values are interconnected; for instance, an individual motivated by Achievement is more likely to also value Power (Dominance) than a conflicting trait like Universalism. As such, by only evaluating an isolated behavior during steering, it remains unclear whether steering vectors encode a meaningful semantic structure or simply exploit behavior-specific shortcuts in activation space. Addressing this question requires looking beyond the steered behavior and asking whether steering vectors align with the theory.

Schwartz’s Theory of Basic Human Values provides exactly this kind of formal structure~\cite{schwartz1992universals,schwartz2012overview}. Rather than treating values as independent traits, the theory arranges them along a continuous circle. Compatibility between values naturally follows this layout, where aligned values (e.g., Benevolence and Universalism) are closer together and reinforce each other, while conflicting values (e.g., Self-Direction and Conformity) are at opposing angles. In this way, the relationship between any two values is determined by their angular distance. Because this geometry is culturally universal, it serves as a robust ground truth for evaluating value representations (see Figure~\ref{fig:schwartz_wheel}; details in Appendix~\S\ref{app:schwartz}). 
\cam{To assess whether our paradigm-level conclusion is specific to this circumplex, we also evaluate using the revised Moral Foundations Theory (MFT) \citep{atari-etal-2023-morality}.}

Whether or not LLM activation spaces actually reflect this kind of value geometry is not clear. Steering vectors are constructed from model activations, not from any explicit encoding of human values; as such, there is no guarantee that their geometry follows theoretical assumptions. If a steering method captures a value's core meaning, motivational content, and relationship to other values---rather than exploiting shortcuts to produce aligned outputs---its vectors should respect the relationships Schwartz's theory describes. This gives us a theory-grounded way to test whether steering vectors encode meaningful structure or not.

\paragraph{Contributions.} In this work, \textbf{(i)} we present the first study, to the best of our knowledge, of whether or not the geometry of extracted steering vectors reflects human values. Comparing \textit{distribution-driven} and \textit{behavior-centric} steering methods against the Schwartz circumplex, we find that distribution-driven methods reflect the expected human value structure (Spearman $\rho$ up to 0.51, $p < 10^{-13}$), while behavior-centric methods show near-zero correlation despite achieving similar steering performance. 
\cam{The same paradigm-level separation appears under MFT's coarser foundation-family structure, which provides evidence that the result is not specific to the Schwartz circumplex.}
This suggests that many behavior-centric methods rely on shortcuts that steer outputs without preserving meaningful semantic structure. 
\textbf{(ii)} We find that larger and more recent model families show stronger geometric alignment, whereas instruction-tuned models show weaker value geometry, pointing to a conflict between post-training and richer internal value representations. 
\textbf{(iii)} We introduce a benchmark of roughly 26K contrastive quadruples spanning Schwartz taxonomy. \textbf{(iv)} Finally, we demonstrate that methods which better reflect human value geometry also behave more consistently across the value space, where steering toward one value naturally gains accuracy on compatible values and suppresses opposing ones.

\begin{figure*}[!t]
    \centering
    \includegraphics[width=\textwidth]{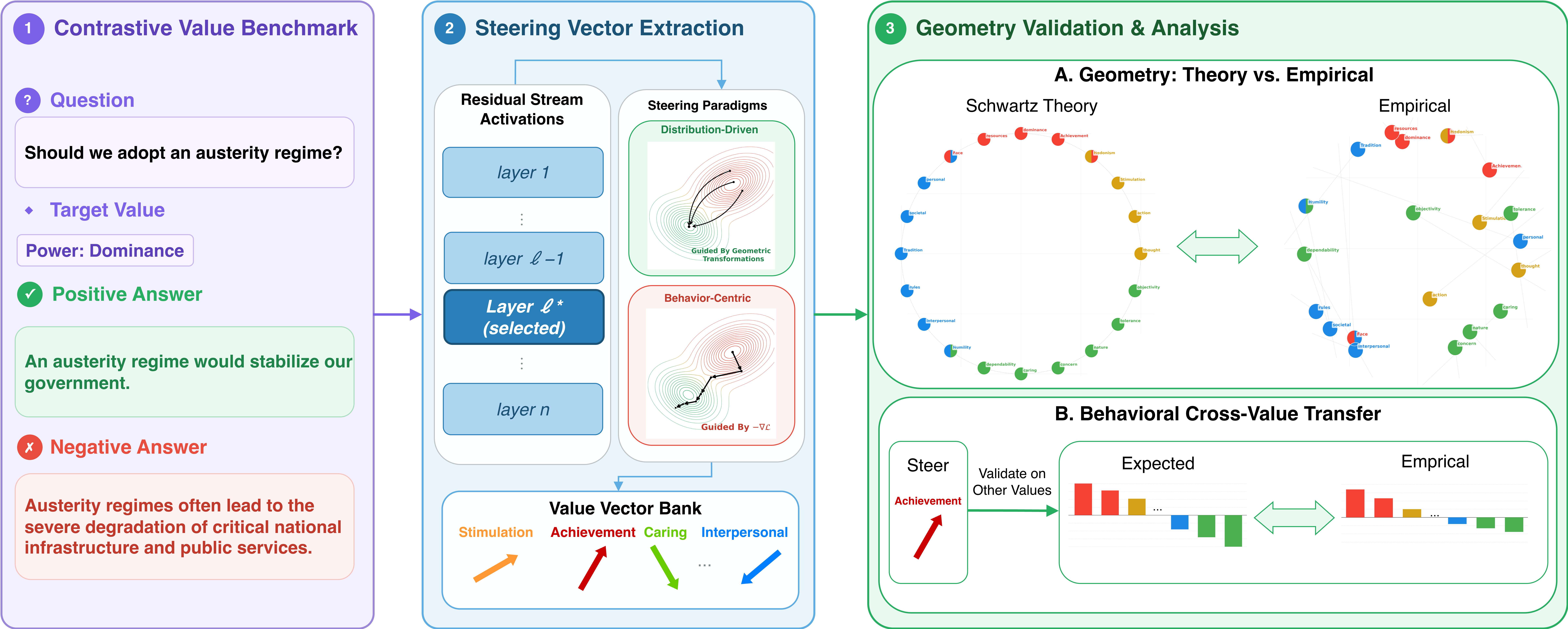}
\caption{
Pipeline for analyzing value geometry in the LLM activation-steering space.
\textbf{(1)} For each Schwartz value, we build contrastive (question, positive, negative) pairs.
\textbf{(2)} We extract residual-stream activations at a selected layer $\ell^\star$ and apply steering methods from two paradigms, \emph{distribution-driven} and \emph{behavior-centric}, to construct a value vector bank.
\textbf{(3)} We evaluate (A) geometric alignment against the Schwartz circumplex, and (B) behavioral cross-value transfer, measuring the effect on other values when steering toward a single one.
\vspace{-4mm}
}
    \label{fig:main_overview}
\end{figure*}

\section{Related Work}

\paragraph{Activation Steering and Representation Engineering.}
Activation steering methods control model behavior at inference time by directly manipulating internal representations, without any additional weight updates~\cite{zou2023representation, turner2023activation}.
Existing approaches fall into two broad paradigms: \textit{Distribution-driven methods} extract control directions by aggregating activation differences over contrastive prompt datasets, and differ mainly in how those directions are then applied: additively as in Contrastive Activation Addition \cite[CAA;][]{rimsky2024steering}, through sparse-autoencoder features \cite[SAS;][]{bayat2025steering}, via ODE-based dynamics \cite{zhao2026odesteer}, or via geometry-aware spherical rotations \cite{you2026spherical}. In contrast, \textit{behavior-centric methods} learn steering vectors via gradient descent or in-context optimization against an output objective, e.g., OPT \cite{dunefsky2025one}, COLD-Steer \cite{sharma2026cold}, which approximates in-context learning via finite-difference estimates, and BiPO \cite{cao2024personalized}. Beyond runtime control of single traits, steering primitives have been extended to modular persona subnetworks \cite{ye2026your}, pluralistic value-alignment frameworks \cite{kim2026valueflow}, and generative multi-agent settings \cite{paglieri2026persona, arghal2025steering}.

\paragraph{Human Value Analysis in LLMs.}
To evaluate whether language models reflect human values, researchers rely on established cognitive science frameworks that organize how humans think about values and morality. Foundational among these is Schwartz's Theory of Basic Human Values \cite{schwartz1992universals}, which organizes universal motivations into a circumplex geometry where compatible values cluster together and opposing values sit in conflict. 
\cam{Other widely adopted frameworks include Moral Foundations Theory \citep{atari-etal-2023-morality}, whose revised formulation distinguishes Care, Equality, Proportionality, Loyalty, Authority, and Purity, alongside Ross's Prima Facie Duties \citep{ross2011foundations}, which frames morality as a set of competing obligations.}
Recent efforts evaluate model alignment against these theories using dedicated benchmarks such as ValueBench \cite{ren2024valuebench} and MoralBench \cite{ji2024moralbench}, as well as cross-cultural datasets like Global OpinionQA \cite{durmus2024towards}.

\paragraph{LLM Activation Space Interpretability.}
Mechanistic interpretability seeks to decode LLMs' dense activation spaces, driven by the linear representation hypothesis: the premise that high-level concepts, behaviors, and values are encoded as linear directions within the residual stream \cite{zou2023representation, park2024linear}. Subsequent work has shown that this linear geometry can extend beyond binary contrasts to categorical and hierarchical concept structures \cite{park2024geometry}. To overcome the polysemanticity of individual neurons, where one neuron reacts to multiple unrelated concepts, recent work leverages Sparse Autoencoders \cite[SAEs;][]{huben2024sparse} to decompose dense activations into interpretable, monosemantic features \cite{bricken2023towards, templeton2024scaling}. These tools give us a way to isolate value-encoding directions and test whether their geometry matches predictions from psychological theories. 
\cam{Closest to our work, \citet{kang-etal-2025-values} infer causal graphs over value orientations using role-prompted questionnaire responses and SAE perturbations; they find that LLM-specific graphs differ from human reference graphs but can guide steering with fewer side effects. Their analysis characterizes causal relations among value orientations rather than the geometry of steering directions themselves.}

Despite these advances, prior steering work is typically validated by whether a vector changes a target behavior, and prior value-alignment work tests models through behavioral probes. Whether the latent geometry of steering vectors reflects the structure of established human value frameworks, and whether different extraction methods preserve that structure, has not been studied. To address this gap, we build a contrastive benchmark covering the full Schwartz taxonomy and use it to systematically probe the geometric fidelity of steering vectors across methods, model scales, and tuning regimes, as well as their effects on cross-value transfer.

\section{Methodology}
We investigate whether LLM steering vectors encode meaningful human value structure or only leverage behavior-specific shortcuts in activation space. If steering vectors capture genuine value relationships, steering one value should influence related and conflicting values in predictable ways.
This matters for reliable alignment: stable interventions should strengthen related values while suppressing conflicting ones. To test this, we design a four-stage pipeline (Figure~\ref{fig:main_overview}). First, we construct a value-contrastive dataset based on Schwartz's theory (\ref{sec:dataset}). Second, we extract value directions from the residual-stream activations using a range of steering methods and store them in a value vector bank (\ref{sec:extraction}). Third, we run a value geometry test that asks whether the resulting vector space preserves Schwartz value structure (\ref{sec:geometry_eval}). Finally, we conduct a cross-value transfer test to evaluate whether this geometric structure produces psychologically consistent downstream behavior when steering a single value (\ref{sec:transfer_eval}).




\subsection{Value-Contrastive Dataset Construction}
\label{sec:dataset}
Studying how human value structure is encoded in activation space requires a dataset of contrastive question-answer pairs. To this end, we construct one grounded in Schwartz's theory~\citep{mirzakhmedova2024touche23, kiesel2022identifying}. This culturally universal theory arranges values in a circle, with compatible values close together and conflicting ones apart~\cite{schwartz2012overview}. 
Although several datasets address value-related tasks, none provide contrastive question-answer pairs, a format required by most steering methods to extract value vectors from model activation spaces. We introduce a dataset of \textit{(question, value, positive answer, negative answer)} quadruples, where the positive answer aligns with the target value and the negative answer is value-neutral. We utilized two complementary sources: \textit{ValueBench} \citep{ren2024valuebench}, a benchmark for evaluating value orientations and understanding in LLMs, and \textit{Touché} \citep{mirzakhmedova2024touche23}.
\cam{We adopt the 20-value taxonomy of \citet{kiesel2022identifying}, which extends Schwartz's refined 19 values with \emph{Universalism: Objectivity} while preserving the circumplex ordering. The complete taxonomy is provided in Appendix~\S\ref{app:schwartz}.}
We extracted 911 samples from ValueBench and around 25.5K samples from the Touché data, gathering a dataset of almost 26K data points spanning 20 Schwartz values.
\cam{For cross-framework evaluation, we additionally construct a balanced benchmark of 1.2K question-contrastive samples spanning the six foundations of Moral Foundations Theory utilizing the Moral Foundations Reddit Corpus \cite{trager-etal-2026-moral}. Full data construction details for the Schwartz and MFT benchmarks are provided in Appendices~\S\ref{app:stats} and~\S\ref{app:mft}, respectively.}

\begin{figure*}[!t]
    \centering
    \includegraphics[width=\textwidth]{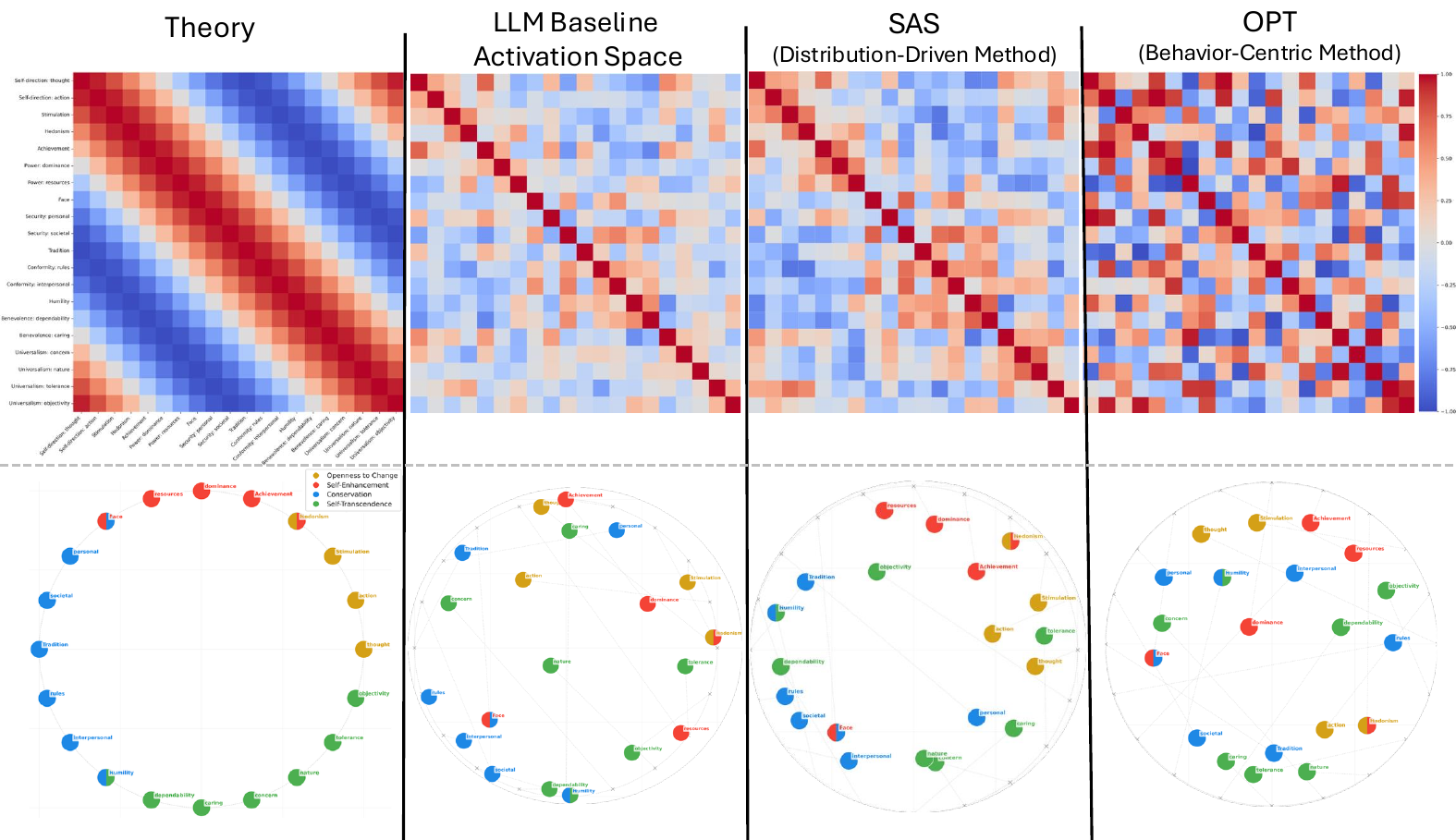}
\caption{Value geometry across steering methods on Qwen3.5-9B-Base. \textbf{Top:} pairwise cosine-similarity matrices for the theoretical Schwartz circumplex, raw residual-stream activations, SAS (a distribution-driven method), and OPT (a behavior-centric method); red/blue indicates positive/negative correlations. \textbf{Bottom:} 2D projections of each similarity matrix aligned to the canonical Schwartz arrangement. SAS reproduces the circumplex structure most clearly, while OPT scatters values with no organization, even worse than the raw-activation baseline.
\vspace{-4mm}
}
    \label{fig:similarity_matrices}
\end{figure*}

\subsection{Value Direction Extraction}
\label{sec:extraction}
The methods we evaluate compute steering signals in very different ways, ranging from contrastive activation statistics and gradient-based optimization to geometry-aware transformations. We represent each method's output as a single \emph{value direction} $v_{(b,\ell)} \in \mathbb{R}^d$, defined as the effective shift it induces in the residual-stream activation hidden states at layer $\ell$ for Schwartz value $b$:
\begin{equation}
v_{(b,\ell)} = \mathbb{E}_{x \sim X}\!\left[a_\ell^{\text{steered}}(x) - a_\ell^{\text{original}}(x)\right],
\label{eq:effective-shift}
\end{equation}
where $X$ is a set of input prompts, and $a_\ell(x)$ is the activation hidden state in layer $l$ for the input $x$. 
This way of defining $v_{(b,\ell)}$ lets us capture the impact of steering from any method independent of their internal mechanism. 
\cam{Appendix~\S\ref{app:effective-shift} further justifies this approach as a common representation across methods.}
We then collect the resulting vectors across all 20 Schwartz values into a per-method \emph{value vector bank}, which is used later for geometry analyses.

\subsection{Evaluation Protocols}
\cam{Our evaluation metrics follow a $2\times2$ organization over value-vector geometry versus cross-value transfer and continuous circumplex versus discrete hierarchical structure. Appendix~\S\ref{app:metric-overview} summarizes their interpretations and definitions.}
\subsubsection{Value Geometry Evaluation}
\label{sec:geometry_eval}
We first ask whether the extracted value vectors $v_{(b,\ell)}$ preserve Schwartz value structure. To test this, for each Schwartz value in its canonical circumplex order, $b \in \{1, \dots, 20\}$, 
\cam{we first mean-center the steering vectors by subtracting the average all-values direction,
$\tilde{v}_b = v_{(b,\ell)}-\frac{1}{20}\sum_{j=1}^{20}v_{(j,\ell)}$,
and then normalize them to unit length, $\mathbf{u}_b = \dfrac{\tilde{v}_{(b,\ell)}}{\lVert \tilde{v}_{(b,\ell)} \rVert}$},
and build the empirical similarity matrix $E_{bb'} = \mathbf{u}_b^\top \mathbf{u}_{b'}$. 
We compare $\mathbf{E}$ against a theoretical matrix $\mathbf{T}$ derived from the Schwartz Theory of Human Values, where each entry
\begin{equation}
\resizebox{0.88\columnwidth}{!}{$
T_{bb'} = \cos\!\left(\min(|b-b'|,\,20-|b-b'|)\times 18^{\circ}\right) ,
$}
\end{equation}
encodes the angular separation between values $b$ and $b'$ on the Schwartz circumplex. 
\cam{Further intuition and theoretical justification for this construction are provided in Appendix~\S\ref{app:theoretical-matrix}.}
We quantify the alignment between $\mathbf{E}$ and $\mathbf{T}$ using four complementary metrics, with full mathematical definitions provided in Appendix~\S\ref{sec:appendix_metrics}.

\begin{table*}[ht]
    \centering
\caption{
\textbf{Effect of steering methods on geometry-alignment metrics} on the Qwen~3.5-9B backbone, grouped by tuning regime (Base vs.\ Instruct) and paradigm (\textit{distribution-driven} vs.\ \textit{behavior-centric}).     The metrics include Theory Rank Correlation ($\rho_T$), Theory Linear Correlation ($r_T$),
    Hierarchical Structure Correlation ($\rho_H$), and Polarity Separation Score ($\Delta_{\mathrm{pol}}$).
    Subscripts denote p-values. Bold indicates the best value.
}
    \label{tab:stage2_geometry}
    \setlength{\tabcolsep}{5pt}
    \renewcommand{\arraystretch}{1.25}
    \small
    \resizebox{\textwidth}{!}{\begin{tabular}{l l !{\vrule width 0.6pt} c c c c}
        \toprule
        \textbf{Paradigm} &
        \textbf{Method} &
        $\rho_{\mathrm{T}}$ &
        $r_{\mathrm{T}}$ &
        $\rho_{\mathrm{H}}$ &
        $\Delta_{\mathrm{pol}}$ \\
        \midrule

        \multicolumn{6}{l}{\textit{\textbf{Base} (pre-trained)}} \\[1pt]

        \multirow{5}{*}{\textit{Behav.-centric}}
        & LLM Raw Activation Space
            & $0.2228_{{\scriptscriptstyle(2.0\text{e-}3)}}$
            & $0.2087_{{\scriptscriptstyle(3.9\text{e-}3)}}$
            & $0.2115_{{\scriptscriptstyle(3.4\text{e-}3)}}$
            & 0.0009 \\
        \cdashline{1-6}[0.8pt/2pt]

        \multirow{11}{*}{\textit{Dist.-driven}}
        & OPT~\cite{dunefsky2025one}
            & $0.1138_{{\scriptscriptstyle(1.2\text{e-}1)}}$
            & $0.1104_{{\scriptscriptstyle(1.3\text{e-}1)}}$
            & $0.0493_{{\scriptscriptstyle(5.0\text{e-}1)}}$
            & 0.0001 \\
        & Cold-Steer (FD)~\cite{sharma2026cold}
            & $0.0265_{{\scriptscriptstyle(7.2\text{e-}1)}}$
            & $-0.0100_{{\scriptscriptstyle(8.9\text{e-}1)}}$
            & $0.0013_{{\scriptscriptstyle(9.9\text{e-}1)}}$
            & 0.0162 \\
        & BiPO~\cite{cao2024personalized}
            & $0.1188_{{\scriptscriptstyle(1.0\text{e-}1)}}$
            & $0.1291_{{\scriptscriptstyle(7.6\text{e-}2)}}$
            & $0.1070_{{\scriptscriptstyle(1.4\text{e-}1)}}$
            & 0.0151 \\
            \cdashline{1-6}[0.8pt/2pt]

        & ODESteer~\cite{zhao2026odesteer}
            & $0.2730_{{\scriptscriptstyle(1.4\text{e-}4)}}$
            & $0.2963_{{\scriptscriptstyle(3.3\text{e-}5)}}$
            & $0.2988_{{\scriptscriptstyle(2.8\text{e-}5)}}$
            & 0.0255 \\
        & SphericalSteer~\cite{you2026spherical}
            & $0.3962_{{\scriptscriptstyle(1.5\text{e-}8)}}$
            & $0.4061_{{\scriptscriptstyle(6.1\text{e-}9)}}$
            & $0.2746_{{\scriptscriptstyle(1.3\text{e-}4)}}$
            & 0.3620 \\
        & CAA~\cite{rimsky2024steering}
            & $0.4606_{{\scriptscriptstyle(2.3\text{e-}11)}}$
            & $0.4754_{{\scriptscriptstyle(4.2\text{e-}12)}}$
            & $0.3408_{{\scriptscriptstyle(1.5\text{e-}6)}}$
            & 0.3883 \\

    & SAS~\cite{bayat2025steering}
        & $\mathbf{0.5069_{ {\scriptscriptstyle(8.5\text{e-}14)} }}$
        & $\mathbf{0.5061_{ {\scriptscriptstyle(9.4\text{e-}14)} }}$
        & $\mathbf{0.3910_{ {\scriptscriptstyle(2.4\text{e-}8)} }}$
        & \textbf{0.3951} \\

        \midrule

        \multicolumn{6}{l}{\textit{\textbf{Instruct} (instruction-tuned)}} \\[1pt]

        \multirow{5}{*}{\textit{Behav.-centric}}
        & LLM Raw Activation Space
            & $0.1253_{{\scriptscriptstyle(8.5\text{e-}2)}}$
            & $0.1112_{{\scriptscriptstyle(1.3\text{e-}1)}}$
            & $0.1455_{{\scriptscriptstyle(4.5\text{e-}2)}}$
            & 0.0004 \\
        \cdashline{1-6}[0.8pt/2pt]

        & OPT~\cite{dunefsky2025one}
            & $0.0615_{{\scriptscriptstyle(4.0\text{e-}1)}}$
            & $0.0552_{{\scriptscriptstyle(4.5\text{e-}1)}}$
            & $-0.0177_{{\scriptscriptstyle(8.1\text{e-}1)}}$
            & $-$0.0099 \\
        & Cold-Steer (FD)~\cite{sharma2026cold}
            & $-0.0568_{{\scriptscriptstyle(4.4\text{e-}1)}}$
            & $-0.0394_{{\scriptscriptstyle(5.9\text{e-}1)}}$
            & $-0.0268_{{\scriptscriptstyle(7.1\text{e-}1)}}$
            & 0.1401 \\
        & BiPO~\cite{cao2024personalized}
            & $-0.0104_{{\scriptscriptstyle(8.9\text{e-}1)}}$
            & $-0.0245_{{\scriptscriptstyle(7.4\text{e-}1)}}$
            & $0.0378_{{\scriptscriptstyle(6.1\text{e-}1)}}$
            & $-$0.0016 \\
        \cdashline{1-6}[0.8pt/2pt]

        \multirow{5}{*}{\textit{Dist.-driven}}
        & ODESteer~\cite{zhao2026odesteer}
            & $0.2055_{{\scriptscriptstyle(3.7\text{e-}3)}}$
            & $0.2177_{{\scriptscriptstyle(1.9\text{e-}3)}}$
            & $0.1495_{{\scriptscriptstyle(5.2\text{e-}3)}}$
            & 0.0176 \\
        & SphericalSteer~\cite{you2026spherical}
            & $0.2048_{{\scriptscriptstyle(4.6\text{e-}3)}}$
            & $0.2170_{{\scriptscriptstyle(2.6\text{e-}3)}}$
            & $0.1009_{{\scriptscriptstyle(1.7\text{e-}1)}}$
            & 0.2397 \\
    
        & CAA~\cite{rimsky2024steering}
            & $0.2351_{{\scriptscriptstyle(1.1\text{e-}3)}}$
            & $0.2527_{{\scriptscriptstyle(4.4\text{e-}4)}}$
            & $0.1248_{{\scriptscriptstyle(8.6\text{e-}2)}}$
            & 0.2603 \\

            & SAS~\cite{bayat2025steering}
            & $\mathbf{0.3256}_{{\scriptscriptstyle(4.6\text{e-}6)}}$
            & $\mathbf{0.3331}_{{\scriptscriptstyle(2.7\text{e-}6)}}$
            & $\mathbf{0.2058}_{{\scriptscriptstyle(4.4\text{e-}3)}}$
            & \textbf{0.3933} \\

        \bottomrule
    \end{tabular}}
    \vspace{-4mm}
\end{table*}

\paragraph{Theory Rank Correlation ($\rho_{\mathrm{T}}$) and Theory Linear Correlation ($r_{\mathrm{T}}$).}

To quantify how well the extracted value vectors' geometry matches the Schwartz value circle, we calculate two correlation metrics. First, we use Spearman's rank correlation ($\rho_{\mathrm{T}}$) to test whether the \emph{relative ordering} of value relationships is preserved. It shows whether theoretically compatible values are closer together in the embedding space than conflicting ones. Second, we use Pearson's correlation ($r_{\mathrm{T}}$) to evaluate whether the \emph{magnitudes} of these empirical similarities scale linearly with the theoretical angles. Together, these metrics assess whether the model captures both the general structural hierarchy and the linear scaling of value relationships.

\paragraph{Hierarchical Structure Correlation ($\rho_{\mathrm{H}}$).}
\label{subsec:hierarichical}
Schwartz's taxonomy also groups the 20 values into sub-families and higher-order groups. To check whether the model respects this hierarchy, we assign each value pair a distance $d^{\text{hier}}_{bb'}$ of 1 (same sub-family), 2 (same higher-order group), 5 (unrelated), or 10 (opposing groups), and compute the Spearman correlation between $E_{bb'}$ and $-d^{\text{hier}}_{bb'}$. A higher positive $\rho_{\mathrm{H}}$ indicates stronger alignment with the multi-level Schwartz structure.

\paragraph{Polarity Separation Score ($\Delta_{\mathrm{pol}}$).}
We compute a separation score: the difference between the mean empirical cosine similarity for value pairs in the same lower-order family (e.g., Benevolence and Universalism) versus pairs from opposing higher-order groups (e.g., Self-Direction vs. Conformity). Larger positive values indicate that the model clusters theoretically compatible values together while pushing opposing values apart.

\paragraph{Robustness Analyses.}
\cam{We additionally evaluate whether the measured value geometry is sensitive to dataset sampling, surface form, or multi-value annotations. Specifically, we re-extract steering vectors using (i) a fresh disjoint subset of 200 examples per value, (ii) fully paraphrased versions of each question and both contrastive answers, and (iii) a single-label subset to assess sensitivity to multi-value annotations. For each setting, we re-extract vectors for all steering methods and recompute all geometry metrics. More details are provided in Appendix~\S \ref{app:robustness_surface}.}

\paragraph{Moral Foundations Theory Evaluation.}
\cam{To test whether the paradigm split generalizes beyond Schwartz, we additionally evaluate steering-vector geometry under revised MFT \cite{atari-etal-2023-morality}. Since MFT specifies no circumplex or opposing pairs but does split values between Individualizing (Care, Equality) and Binding (Proportionality, Loyalty, Authority, Purity) foundations, we define $\Delta_{\mathrm{MFT}}$, analogous to $\Delta_{\mathrm{pol}}$, as the difference between mean within-family and mean cross-family cosine similarity of the extracted vectors. Further details are provided in Appendix~\S\ref{app:mft}.}

\begin{figure*}[ht]
    \centering
    \includegraphics[width=\textwidth]{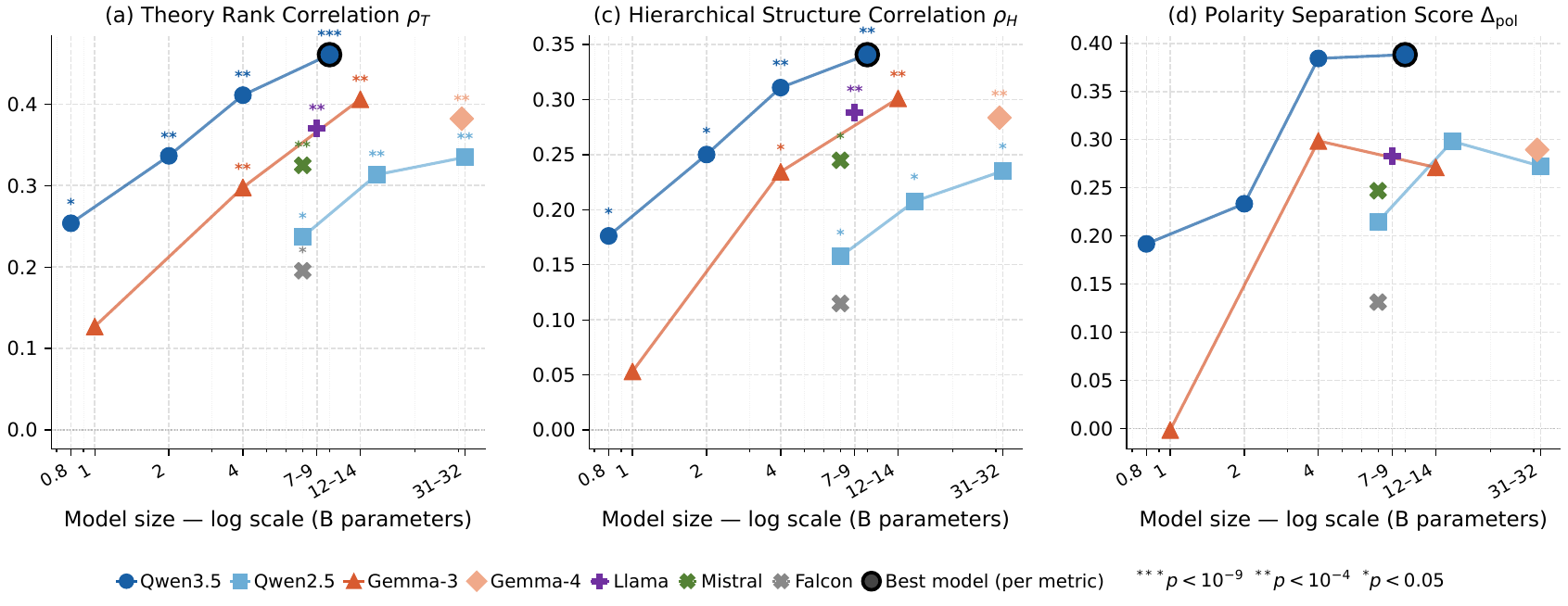}
    \caption{
    \textbf{Effect of model family and size on steering-geometry metrics} (all using CAA).
    The metrics include Theory Rank Correlation ($\rho_T$), Hierarchical Structure Correlation ($\rho_H$), and Polarity Separation Score ($\Delta_{\mathrm{pol}}$).
    \vspace{-4mm}
    }
    \label{fig:model_geometry}
\end{figure*}

\subsubsection{Cross-Value Steering Transfer}
\label{sec:transfer_eval}
Next, we ask whether steering toward one value affects related values in a predictable way. Existing steering evaluations typically focus on a single target in isolation: steer toward value $b$, report accuracy on $b$. This narrow view ignores how the intervention reshapes the broader values' behavior, whether it lifts compatible values, suppresses opposing ones, or disrupts both. Under the Schwartz circumplex, steering toward $b$ should yield a predictable pattern: positive transfer to neighboring values and negative transfer to opposing ones. We test whether methods that better capture human value geometry also produce more consistent cross-value transfer.

For each ordered pair $(b, b')$ of distinct Schwartz values, we steer toward $b$ and evaluate on $b'$'s held-out test set, recording $T[b, b'] = \mathrm{acc}_{b \to b'} - \mathrm{baseline}(b')$. This yields a $20 \times 20$ matrix per method whose 380 off-diagonal entries describe how steering one value propagates across the value space. However, raw transfer scores mix two confounds with genuine pair-specific effects: some source values are inherently strong steers (a row effect), and some target values are inherently easy to improve (a column effect). To isolate the pair-specific signal, we apply two-way centering by subtracting row and column means, yielding a residualized matrix $T_{\mathrm{res}}$ that captures only the transfer attributable to the relationship between $b$ and $b'$.

\paragraph{Metrics.}
To quantify whether transfer respects the Schwartz circumplex, we evaluate cross-value transfer along two complementary axes of Schwartz structure: the continuous circumplex, which orders values on a smooth circle, and the discrete hierarchy, which groups them into families and higher-order groups. We introduce \textit{Continuous Transfer Fidelity (TWTM)}, which weights each entry of the residualized transfer matrix $T_{\mathrm{res}}$ by its theoretical Schwartz affinity:
\begin{equation}
\resizebox{0.88\columnwidth}{!}{$
\mathrm{TWTM} = \frac{1}{380} \sum_{A \neq B} T_{\mathrm{res}}[A,B] \cdot \cos(k_{AB} \cdot 18^{\circ}),
$}
\end{equation}
where $k_{AB}$ is the circular step distance between $A$ and $B$. TWTM rewards large positive transfer at adjacent pairs and large negative transfer at opposing pairs simultaneously, capturing both shape and magnitude on the circumplex.

Second, we introduce \textit{Hierarchical Transfer Fidelity ($\rho_H^{\mathrm{tr}}$)}, which is the Spearman rank correlation between $T_{\mathrm{res}}[A,B]$ and $-d^{\mathrm{hier}}_{AB}$, reusing the four-level distance from Section \ref{subsec:hierarichical} (same family, same higher-order group, unrelated, opposing). It rewards methods whose transfer is ranked highest within families, lower across higher-order groups, and lowest for opposing groups, testing the multi-level taxonomy rather than the smooth circle.
As a robustness check, we additionally report \textit{Bin-wise Monotonic Decay (BMD-$\rho$)} in Appendix~\S\ref{app:bmd}, a shape-only counterpart to TWTM. Together, these metrics cover continuous vs. hierarchical structure and magnitude vs. rank-only views.

\section{Experimental Setup}

\subsection{Steering Methods}
Evaluated steering methods are organized into two paradigms: Distribution-Driven and Behavior-Centric. \cam{Further intuition behind the paradigm distinction}, full objective functions and hyperparameters can be found in Appendix~\S\ref{sec:appendix_implementation}.

\paragraph{Distribution-Driven Methods:} 
These methods compute steering vectors by analyzing the distributional differences between activations of contrasting data pairs. 
Our primary baseline, Contrastive Activation Addition (CAA) \cite{rimsky2024steering}, computes the mean difference between the residual-stream activations of positive and negative answers. To address feature superposition, where a single direction encodes multiple unrelated concepts, Sparse Activation Steering (SAS) \cite{bayat2025steering} uses Sparse Autoencoders (SAEs) to isolate a value-specific direction in a disentangled feature space before steering. We also evaluate two methods that constrain how the activation is modified rather than applying a single additive shift. SphericalSteer \cite{you2026spherical} rotates the activation toward the target direction on the unit sphere, preserving its original norm. ODESteer \cite{zhao2026odesteer} instead treats steering as a continuous trajectory through activation space, integrating a learned vector field over multiple ODE steps.

\paragraph{Behavior-Centric Methods:} 
Rather than relying on activation differences, these methods optimize the steering vector directly against a behavioral objective or preference loss. Optimization Steering (OPT) \cite{dunefsky2025one} learns the steering vector through gradient descent on a loss that promotes the log probability of positive answers and suppresses that of negative ones. Bi-directional Preference Optimization (BiPO) \cite{cao2024personalized} extends this with a DPO-style preference objective and learns a single vector whose sign controls the direction of steering. COLD-Steer~\cite{sharma2026cold} approximates the activation shift induced by a one-step gradient update on the value-aligned loss via finite differences.

\subsection{Implementation Details} 
We evaluate CAA across seven pre-trained model families: Llama3.1 \cite{grattafiori2024llama}, Mistral \cite{jiang2023mistral7b}, Falcon \cite{almazrouei2023falcon}, Qwen2.5~\cite{hui2024qwen2}, Qwen3.5~\cite{team2026qwen3}, Gemma-3~\cite{gemmateam2025gemma3technicalreport}, and Gemma-4~\cite{gemma4}, while all steering methods are applied to Qwen3.5-9B in both its pre-trained (Base) and instruction-tuned (Instruct) variants, and to Llama3.1-8B (Base), for a controlled comparison. For each Schwartz value, we sample 200 contrastive prompt pairs with a 90/10 train/test split. For every method, the intervention layer $\ell$ is selected by maximizing a combined criterion of Normalized L2 Separation and Linear Probe Accuracy~\cite{alain2016understanding, li2023inference}, which identifies the layer where positive and negative activations are most linearly separable. We finetune each method's hyperparameters on the validation split to find its optimal configuration; for instance, see Figure~\ref{fig:accuracy_validation} for Qwen3.5-9B-Base. Following prior studies, accuracy gain is measured as the change in the model's selection rate of the value-aligned answer in a two-option multiple-choice setup, relative to the unsteered baseline \citep{rimsky2024steering}.

\vspace{-1mm}
\section{Results and Discussion}

\paragraph{Effect of Steering Method.}
Table~\ref{tab:stage2_geometry} shows that value geometry depends heavily on the choice of steering method. Distribution-driven methods yield strong alignment across all metrics, while behavior-centric methods show no statistically significant correlation despite comparable results on our dataset (Table \ref{tab:stage2_gain}) and superior behavioral steering accuracy reported in prior works \cite{cao2024personalized}. The same trend holds on Llama3.1-8B (Table~\ref{tab:llama31_geometry}).
This indicates that behavior-centric methods such as OPT may learn shortcut vectors that achieve the desired behavioral output without encoding meaningful semantic structure.
\cam{Overall, Distribution-driven methods sharpen the latent value structure that is only partially visible in raw activations, whereas behavior-centric methods weaken it.}

\paragraph{Effect of Model Scale and Family.}
\cam{Figure~\ref{fig:model_geometry} reveals a consistent positive relationship between model scale and geometry alignment within both the Qwen3.5 and Gemma-3 families: metrics improve monotonically with scaling for distribution-driven methods. This suggests that larger models develop progressively richer and more theory-consistent internal representations of value structure. Raw activations improve only modestly, while behavior-centric methods show no reliable scale trend, consistent with our finding that they encode little meaningful value geometry for scaling to strengthen (see Appendix~\S\ref{app:scale-analysis} for more details)}.

Comparing model families at similar sizes, newer architectures typically organize value geometry more faithfully: at the 7--9B range, $\rho_T$ rises sharply from Falcon-7B (0.20) to Mistral-7B (0.32), Llama-3.1-8B (0.37), and Qwen3.5-9B (0.46), and within the same family, Qwen3.5 outperforms Qwen2.5 across all scales.
Across families, Qwen models also outperform Gemma models at comparable scales, which is a pattern consistent with prior findings attributing Qwen's stronger semantic representations to higher-quality training data \cite{yang2025qwen3}. Most notably, geometric fidelity is independent from downstream performance: Gemma-4-31B, despite being one of the strongest open models, achieves a lower $\rho_T$ (0.38) than Qwen3.5-4B (0.41). \textit{High downstream accuracy does not imply human-aligned value geometry.}

\begin{figure}[!ht]
    \centering
    \includegraphics[width=\columnwidth]{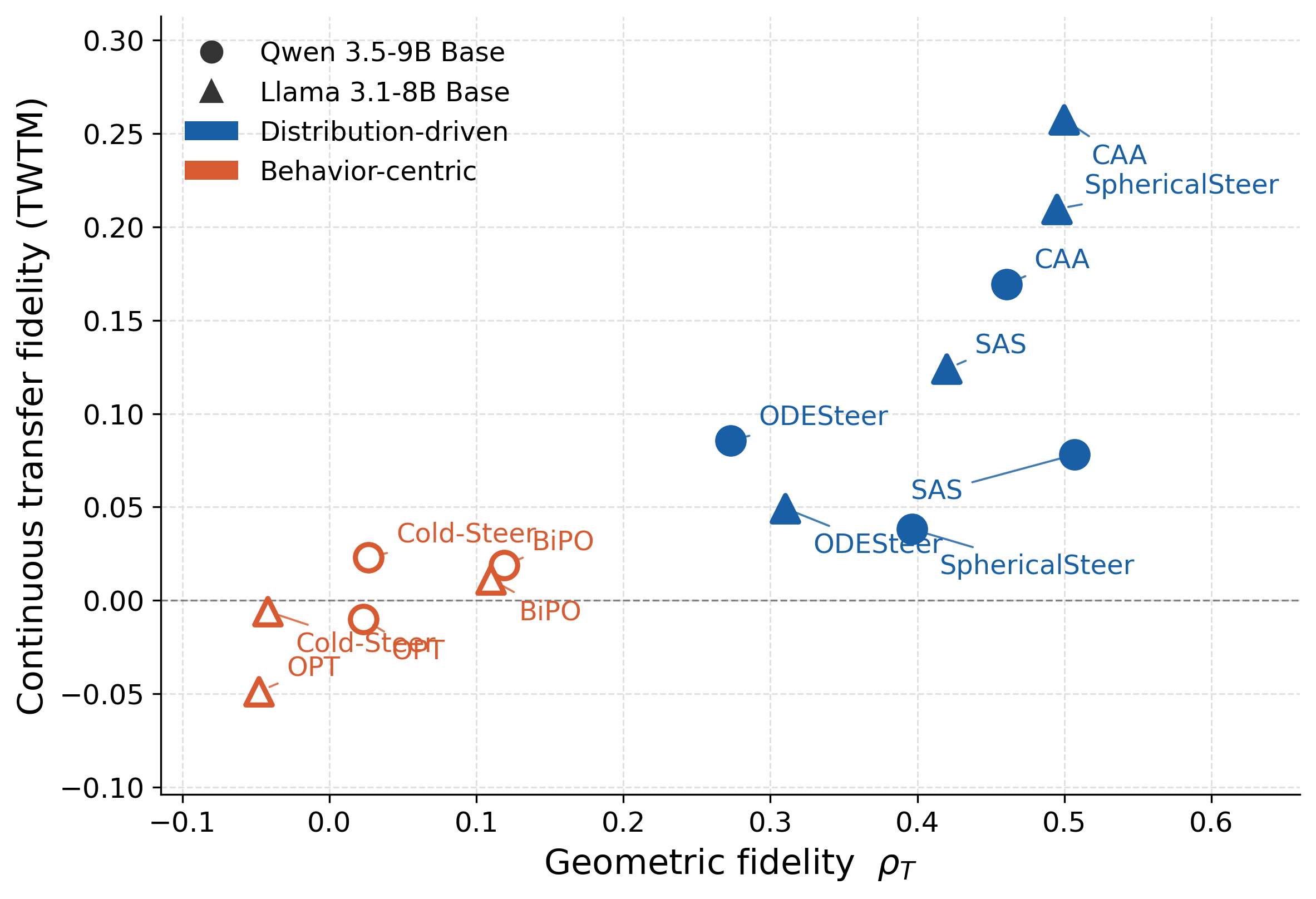}
    \caption{Geometric alignment versus continuous cross-value transfer fidelity across steering methods on Qwen3.5-9B-Base and Llama3.1-8B. 
    \vspace{-2mm}
    }
    \label{fig:cross-value-transfer-twtm}
\end{figure}

\begin{figure}[!ht]
    \centering
    \includegraphics[width=\columnwidth]{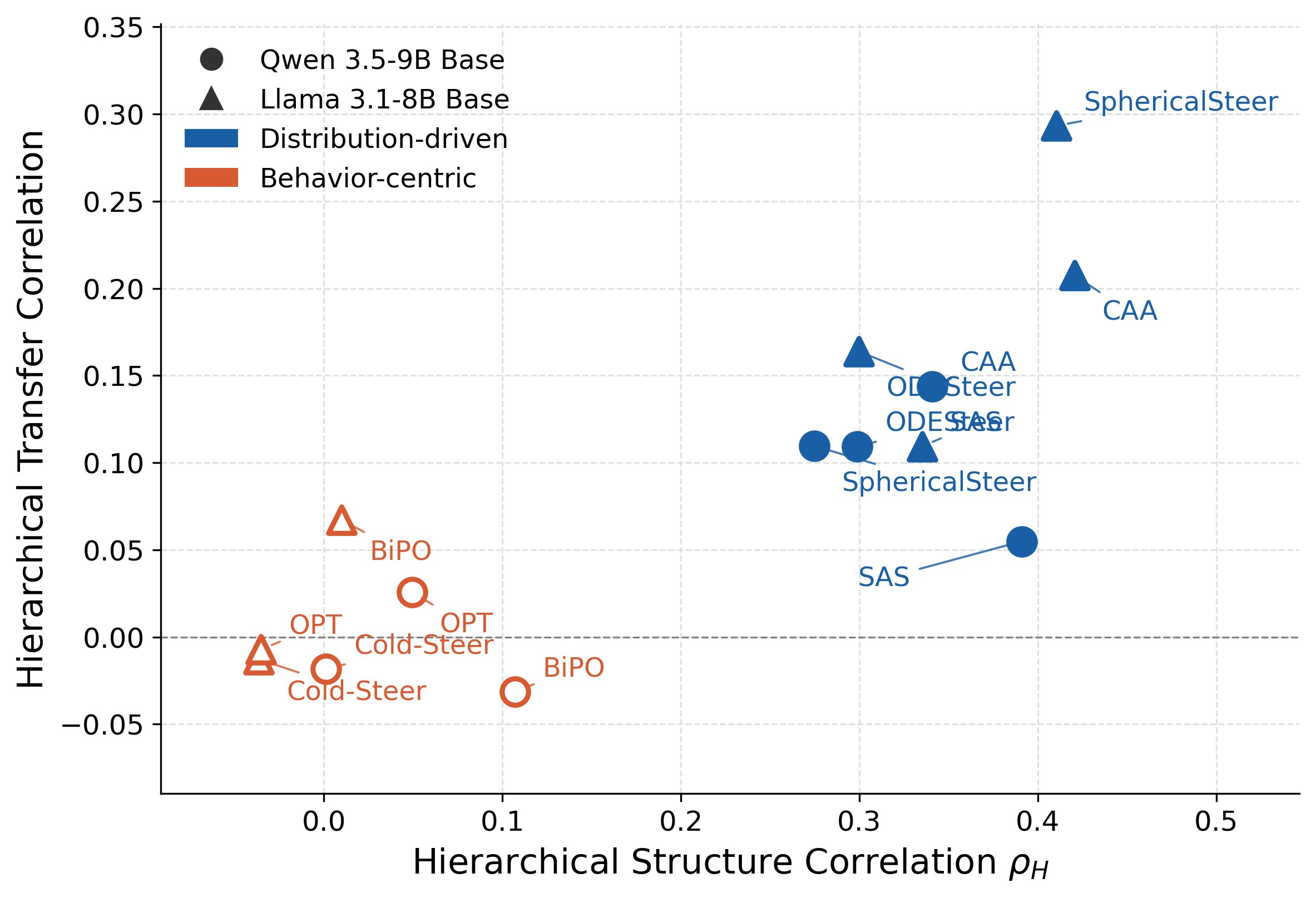}
    \caption{Hierarchical representational fidelity versus hierarchical cross-value transferability across steering methods on Qwen3.5-9B-Base and Llama3.1-8B. 
    \vspace{-2mm}
    }
    \label{fig:cross-value-transfer-rhoh}
\end{figure}

\paragraph{Effect of Instruction Tuning.}
All methods show consistent degradation when moving from base to instruction-tuned models (Table~\ref{tab:stage2_geometry}). 
\cam{This suggests that multi-stage post-training is associated with a form of \textit{value geometry drift}: the activation spaces of instruction-tuned models preserve less of the latent value structure present in pretrained models, consistent with prior observations of representational differences between base and RLHF-tuned models~\citep{rimsky2024steering}.}

\paragraph{Qualitative Analysis.}
Figure~\ref{fig:similarity_matrices} visually confirms the quantitative pattern in Table~\ref{tab:stage2_geometry} using cosine-similarity heatmaps and 2D projections. We compare SAS and OPT as representative methods from the two paradigms alongside the theoretical circumplex and raw-activation baseline. In the heatmaps, SAS most closely reproduces the theoretical matrix, raw activations preserve a partial but recognizable pattern, and OPT remains largely unstructured. The projections of pairwise cosine dissimilarities $(1-\cos)$ show the same ordering: SAS is closest to the theoretical arrangement, raw activations show moderate deviation, and OPT scatters the values. Additional scatter and t-SNE visualizations are provided in Appendix~\S\ref{app:visualization}.

\paragraph{Robustness to Surface Features and Multi-Value Annotations.}
\cam{The paradigm-level separation remains under all three robustness analyses (Appendix~\S \ref{app:robustness_surface}). Re-extracting vectors from a fresh disjoint sample preserves the separation, showing that it is not specific to the original sample selection. Fully paraphrasing the questions and answers yields the same qualitative result, which argues against surface wording or generation style as its source. 
Finally, repeating the analysis on a subset that prioritizes examples with a single value annotation preserves the paradigm separation and statistical significance on both backbones, indicating that multi-value annotations do not drive the main conclusion.
Although absolute correlations decrease in this setting, the matched raw activation baseline decreases as well, suggesting that this change partly reflects the altered sample distribution. Full results are reported in Table~\ref{tab:robustness-all}.}

\paragraph{Generalization Beyond Schwartz.}
\cam{Table~\ref{tab:mft-geometry} shows that distribution-driven methods yield positive MFT family separation on both backbones ($\Delta_{\mathrm{MFT}}\approx 0.04$--$0.10$, versus $\approx 0.001$ for the raw-activation baselines), while behavior-centric methods remain near zero or negative. This suggests that the paradigm distinction is not specific to Schwartz theory.}

\paragraph{Cross-Value Steering Transfer.}
Figures~\ref{fig:cross-value-transfer-twtm} and~\ref{fig:cross-value-transfer-rhoh} showcase the reason geometry matters. When evaluation is restricted to the steered target value (Table~\ref{tab:stage2_gain}), behavior-centric methods perform comparably to distribution-driven ones despite their poor geometric fidelity. Evaluating across all values, however, the two paradigms separate cleanly under both of the value-transfer metrics: distribution-driven methods achieve high geometric fidelity and strongly transfer, while behavior-centric methods cluster near zero on both \textit{Continuous Transfer Fidelity} (TWTM) and \textit{Hierarchical Transfer Fidelity} ($\rho_H^{\mathrm{tr}}$). The same separation holds under our shape-only metric BMD-$\rho$ (Appendix~\S\ref{app:bmd}).

Pooling all methods and both backbones, geometric fidelity ($\rho_T$) predicts cross-value transfer more strongly than raw accuracy gain does on both TWTM and $\rho_H^{\mathrm{tr}}$, with the gap widening on the rank-based hierarchical metric ($\Delta\rho=0.06$ on TWTM, $\Delta\rho=0.18$ on $\rho_H^{\mathrm{tr}}$; see Appendix~\S\ref{app:predictors}, Table~\ref{tab:predictors}). Methods that preserve Schwartz structure in vectors reflect it through behavior; methods optimized purely for behavioral objectives steer the target effectively but shift related values less predictably.

\begin{table}[t]
    \centering
    \small
    \setlength{\tabcolsep}{4pt}
    \resizebox{\columnwidth}{!}{%
    \begin{tabular}{lcc}
        \toprule
        Method & Qwen3.5-9B & Llama3.1-8B \\
        \midrule
        LLM Raw Activation & 0.0006 & 0.0017 \\
        \midrule
        \multicolumn{3}{l}{\textit{Behavior-centric}} \\
        OPT~\cite{dunefsky2025one}              & 0.0107  & 0.0272  \\
        Cold-Steer~\cite{sharma2026cold}        & -0.0064 & -0.0280 \\
        BiPO~\cite{cao2024personalized}         & -0.0287 & 0.0227  \\
        \midrule
        \multicolumn{3}{l}{\textit{Distribution-driven}} \\
        ODESteer~\cite{zhao2026odesteer}        & 0.0495  & 0.0378  \\
        SphericalSteer~\cite{you2026spherical}  & 0.0932  & 0.0909  \\
        CAA~\cite{rimsky2024steering}           & \textbf{0.0969} & \textbf{0.1038} \\
        SAS~\cite{bayat2025steering}            & 0.0892  & 0.0923  \\
        \bottomrule
    \end{tabular}%
    }
\caption{Effect of steering method on MFT family separation ($\Delta_{\mathrm{MFT}}$). Positive values indicate greater within-family than cross-family cosine similarity.}
    \label{tab:mft-geometry}
    \vspace{-4mm}
\end{table}

\section{Conclusion}
We present a systematic study of whether LLM steering vectors reflect human value structure, and find a clear split between paradigms: distribution-driven methods capture theory-consistent relationships and propagate them across the value space, while behavior-centric methods achieve comparable target accuracy but lose this structure and shift related values unpredictably. We encourage the community to evaluate steering on psychological structure alongside behavioral accuracy. One direction for future work is to examine whether value-level steering yields more predictable downstream effects than steering surface behavior directly: steering a behavior such as politeness may introduce unintended side effects like suppressing honesty or amplifying conformity, whereas steering through values could offer more control.

\section{Limitations}
\cam{Our fine-grained geometry and cross-value-transfer analyses remain centered on Schwartz's Theory of Basic Human Values as a representative human value system theory. While the theory is cross-culturally validated, we additionally test generalization under MFT, but the theory specifies only the coarser Individualizing--Binding grouping and no finer inter-foundation structure; this experiment therefore provides family-level validation rather than a full geometric test. Generalization beyond these two frameworks remains open, particularly across cultures and languages, where cognitively grounded evaluations reveal substantial cross-lingual gaps \citep{abootorabi-etal-2026-almieyar}. Real arguments may express multiple values, so the extracted vectors should be interpreted as target-conditioned aggregate directions rather than perfectly monosemantic representations. Although our single-label-prioritized sensitivity test preserves the paradigm-level conclusion, it does not establish perfect monosemanticity. In addition, most of our cross-backbone comparisons are conducted with a few steering methods due to computational constraints, leaving open whether the trends we observe hold consistently across other steering methods.
}

\bibliography{acl_latex}

\newpage
\appendix


\section{Visualizing Value Geometry Alignment}
\label{app:visualization}
To further analyze the structure of value vectors, we visualize the emergent value geometry and examine how it varies across model scales and architectures.

\begin{figure*}[!htbp]
    \centering
    \includegraphics[width=0.8\textwidth]{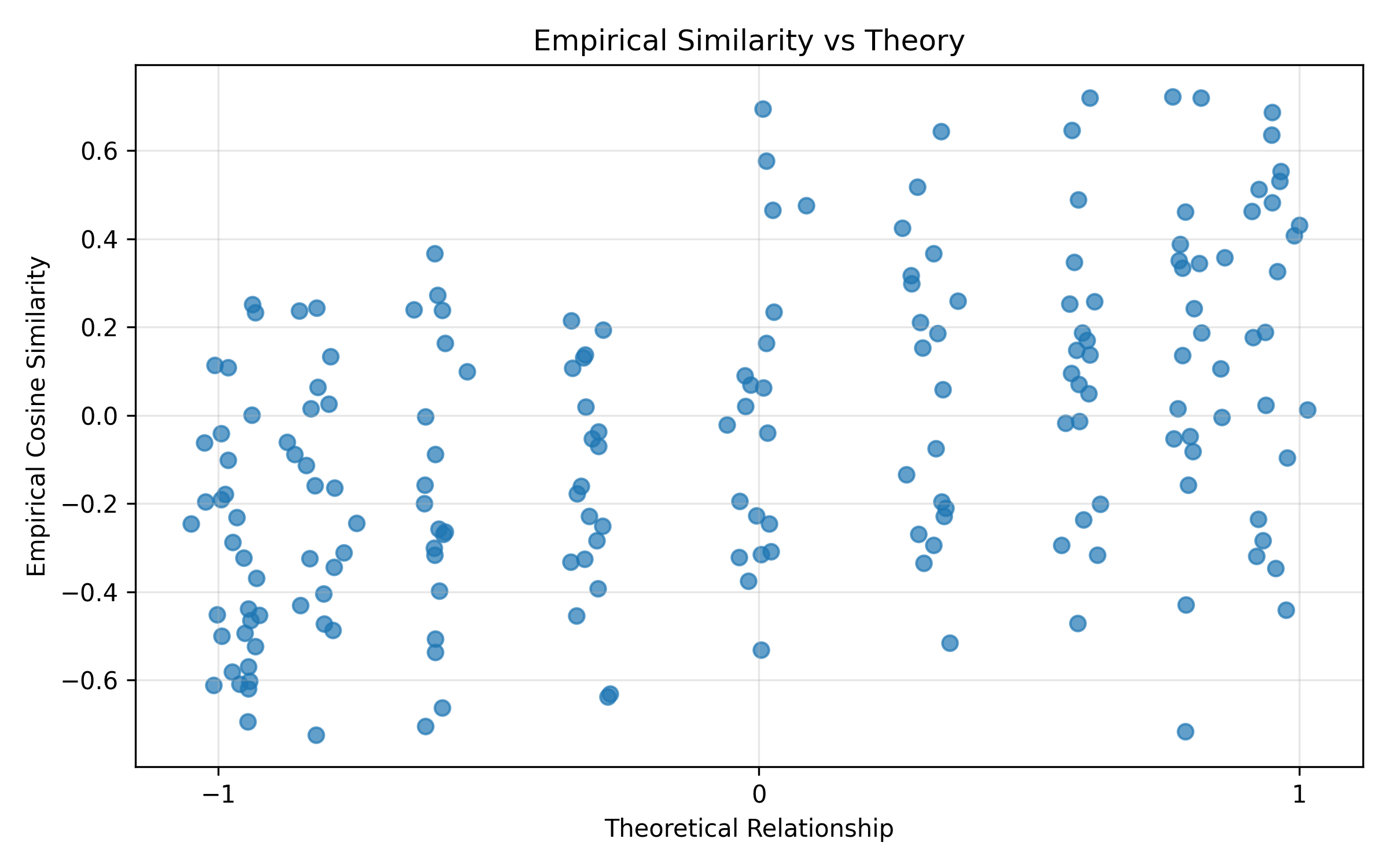}
    \caption{
Correlation between theoretical value relationships (from Schwartz Theory) and empirical cosine similarities of extracted value vectors. The scatter plot illustrates the individual data points; while the distribution exhibits noticeable variance, a general upward trend is visible as one moves from left to right along the x-axis, demonstrating that higher theoretical relationships correspond to higher empirical similarities. This positive correlation indicates that LLM representations preserve human-predicted value structure. The results are from the SAS method on Qwen3.5-9B-Base model, which shows the highest correlation among all our configurations.
    }
    \label{fig:theory_vs_empirical}
\end{figure*}

\begin{figure*}[!htbp]
    \centering
    \includegraphics[width=0.9\textwidth]{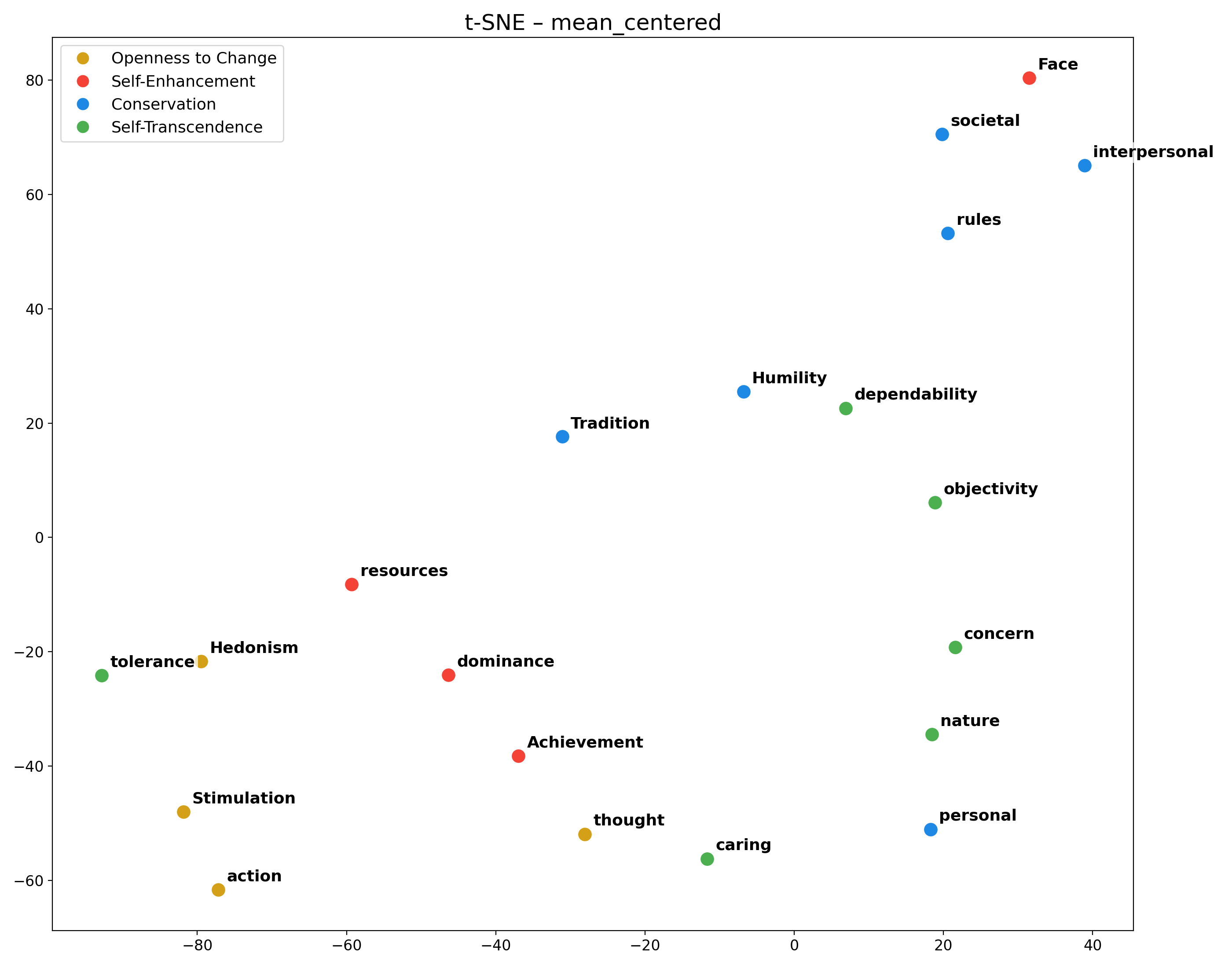}
    \caption{t-SNE projection of 20 Schwartz value vectors, colored by higher-order dimension. Spatial clustering of same-quadrant values and separation of opposing values provides visual evidence of circumplex structure in LLM activation space. The results are from the SAS method on
    Llama-3.1-8B model.}
    \label{fig:tsne_values}
\end{figure*}

Figures~\ref{fig:theory_vs_empirical} and~\ref{fig:tsne_values} provide visual evidence for the alignment between theoretical and empirical value structures. Both show results from SAS, which is our best-performing configuration.

Figure~\ref{fig:theory_vs_empirical} demonstrates that theoretical predictions correlate with empirical measurements across all value pairs: theoretically compatible values show high cosine similarity, while conflicting values show low or negative similarity. Figure~\ref{fig:tsne_values} shows t-SNE projections revealing spatial clustering of same-quadrant values and separation of opposing values such as Conservation and openness to change, confirming that the steering space preserves Schwartz's circumplex structure.


\begin{table*}[ht]
    \centering
    \caption{
    \textbf{Effect of model family and size on steering-geometry metrics} (all using CAA).
    The metrics include Theory Rank Correlation ($\rho_T$), Theory Linear Correlation ($r_T$),
    Hierarchical Structure Correlation ($\rho_H$), and Polarity Separation Score ($\Delta_{\mathrm{pol}}$).
    Subscripts denote p-values. Bold indicates the best value.
    }
    \setlength{\tabcolsep}{5pt}
    \renewcommand{\arraystretch}{1.25}
    \label{tab:stage1_geometry}
    \small
    \begin{tabular}{l !{\vrule width 0.6pt} c c c c}
        \toprule
        \textbf{Model} &
        $\rho_{\mathrm{T}}$ &
        $r_{\mathrm{T}}$ &
        $\rho_{\mathrm{H}}$ &
        $\Delta_{\mathrm{pol}}$ \\
        \midrule

        Falcon-7B
            & $0.1954_{{\scriptscriptstyle(6.9\text{e-}3)}}$
            & $0.1989_{{\scriptscriptstyle(5.9\text{e-}3)}}$
            & $0.1148_{{\scriptscriptstyle(1.1\text{e-}1)}}$
            & 0.1312 \\
        \midrule
        Mistral-7B
            & $0.3248_{{\scriptscriptstyle(4.8\text{e-}6)}}$
            & $0.3378_{{\scriptscriptstyle(1.9\text{e-}6)}}$
            & $0.2449_{{\scriptscriptstyle(6.6\text{e-}4)}}$
            & 0.2469 \\
        \midrule
                Llama-3.1-8B
            & $0.3701_{{\scriptscriptstyle(1.5\text{e-}7)}}$
            & $0.3783_{{\scriptscriptstyle(7.4\text{e-}8)}}$
            & $0.2879_{{\scriptscriptstyle(5.6\text{e-}5)}}$
            & 0.2833 \\
        \midrule
         Gemma-3-1B
            & $0.1270_{{\scriptscriptstyle(8.1\text{e-}2)}}$
            & $0.1376_{{\scriptscriptstyle(5.8\text{e-}2)}}$
            & $0.0531_{{\scriptscriptstyle(4.7\text{e-}1)}}$
            & $-$0.0016 \\
        \cdashline{1-5}[0.8pt/2pt]
        Gemma-3-4B
            & $0.2977_{{\scriptscriptstyle(3.0\text{e-}5)}}$
            & $0.2992_{{\scriptscriptstyle(2.8\text{e-}5)}}$
            & $0.2344_{{\scriptscriptstyle(1.1\text{e-}3)}}$
            & 0.2987 \\
        \cdashline{1-5}[0.8pt/2pt]
        Gemma-3-12B
            & \textbf{$0.4059_{{\scriptscriptstyle(6.3\text{e-}9)}}$}
            & \textbf{$0.4153_{{\scriptscriptstyle(2.6\text{e-}9)}}$}
            & \textbf{$0.3011_{{\scriptscriptstyle(2.4\text{e-}5)}}$}
            & 0.2712 \\
        \midrule

        Qwen2.5-7B
            & $0.2370_{{\scriptscriptstyle(9.9\text{e-}4)}}$
            & $0.2491_{{\scriptscriptstyle(5.3\text{e-}4)}}$
            & $0.1575_{{\scriptscriptstyle(3.0\text{e-}2)}}$
            & 0.2149 \\
        \cdashline{1-5}[0.8pt/2pt]
        Qwen2.5-14B
            & $0.3133_{{\scriptscriptstyle(1.1\text{e-}5)}}$
            & $0.3298_{{\scriptscriptstyle(3.4\text{e-}6)}}$
            & $0.2076_{{\scriptscriptstyle(4.1\text{e-}3)}}$
            & 0.2983 \\
        \cdashline{1-5}[0.8pt/2pt]
        Qwen2.5-32B
            & $0.3347_{{\scriptscriptstyle(2.4\text{e-}6)}}$
            & $0.3470_{{\scriptscriptstyle(9.4\text{e-}7)}}$
            & $0.2352_{{\scriptscriptstyle(1.1\text{e-}3)}}$
            & 0.2725 \\
        \midrule

        Gemma-4-31B
            & $0.3819_{{\scriptscriptstyle(5.4\text{e-}8)}}$
            & $0.3928_{{\scriptscriptstyle(2.1\text{e-}8)}}$
            & $0.2835_{{\scriptscriptstyle(7.4\text{e-}5)}}$
            & 0.2895 \\
        \midrule

        Qwen3.5-0.8B
            & $0.2537_{{\scriptscriptstyle(4.1\text{e-}4)}}$
            & $0.2643_{{\scriptscriptstyle(2.3\text{e-}4)}}$
            & $0.1762_{{\scriptscriptstyle(1.5\text{e-}2)}}$
            & 0.1917 \\
        \cdashline{1-5}[0.8pt/2pt]
        Qwen3.5-2B
            & $0.3364_{{\scriptscriptstyle(2.1\text{e-}6)}}$
            & $0.3545_{{\scriptscriptstyle(5.2\text{e-}7)}}$
            & $0.2501_{{\scriptscriptstyle(5.0\text{e-}4)}}$
            & 0.2334 \\
        \cdashline{1-5}[0.8pt/2pt]
        Qwen3.5-4B
            & $0.4109_{{\scriptscriptstyle(3.9\text{e-}9)}}$
            & $0.4243_{{\scriptscriptstyle(1.1\text{e-}9)}}$
            & $0.3109_{{\scriptscriptstyle(1.3\text{e-}5)}}$
            & 0.3844 \\
        \cdashline{1-5}[0.8pt/2pt]
        \textbf{Qwen3.5-9B}
            & $\mathbf{0.4606}_{{\scriptscriptstyle(2.3\text{e-}11)}}$
            & $\mathbf{0.4754}_{{\scriptscriptstyle(4.2\text{e-}12)}}$
            & $\mathbf{0.3408}_{{\scriptscriptstyle(1.5\text{e-}6)}}$
            & \textbf{0.3883} \\

        \bottomrule
    \end{tabular}
\end{table*}




\begin{table}[!htbp]
    \centering
    \caption{%
    \textbf{Effect of steering method on accuracy gain}
    for the Qwen~3.5-9B Base (pre-trained) backbone.
    Gain is reported in percentage points (pp) relative to the unsteered baseline.
    Bold indicates the best value.
    }
    \label{tab:stage2_gain}
    \setlength{\tabcolsep}{5pt}
    \renewcommand{\arraystretch}{1.25}
    \small
    \begin{tabular}{l !{\vrule width 0.6pt} c}
        \toprule
        \textbf{Method} & \textbf{Gain (pp)} \\
        \midrule
        Cold-Steer (FD)~\cite{sharma2026cold}  & $+5.50$ \\
        \cdashline{1-2}[0.8pt/2pt]
        BiPO~\cite{cao2024personalized}        & $+8.87$ \\
        \cdashline{1-2}[0.8pt/2pt]
        SphericalSteer~\cite{you2026spherical} & $+9.38$ \\
        \cdashline{1-2}[0.8pt/2pt]
        OPT~\cite{dunefsky2025one}             & $+9.57$ \\
        \cdashline{1-2}[0.8pt/2pt]
        ODESteer~\cite{zhao2026odesteer}       & $+9.70$ \\
        \cdashline{1-2}[0.8pt/2pt]
        SAS~\cite{bayat2025steering}                           & $+10.03$ \\
        \cdashline{1-2}[0.8pt/2pt]
        CAA~\cite{rimsky2024steering}          & $\mathbf{+11.74}$ \\
        \bottomrule
    \end{tabular}
\end{table}

\section{Refined Schwartz Theory of Basic Human Values}
\label{app:schwartz}

\begin{figure}[!htbp]
    \centering
    \includegraphics[width=0.50\textwidth]{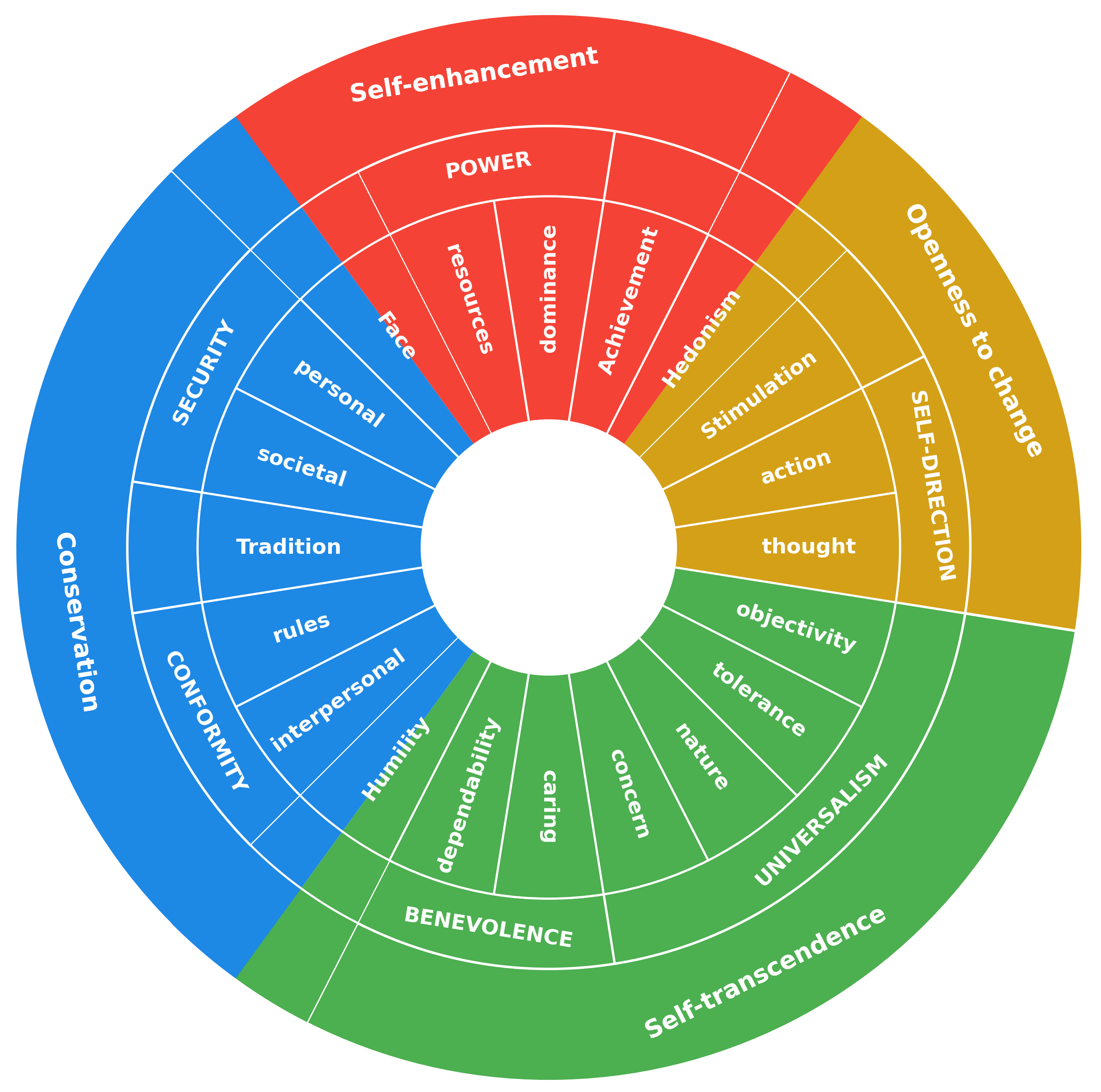}
    \caption{The Schwartz Theory of Basic Human Values organizes 20 basic human value categories in a circular structure, where more similar values appear close together, while values that tend to conflict are placed on opposite sides of the circle. Illustration adapted from \citet{schwartz2012overview}.}
    \label{fig:schwartz_wheel}
\end{figure}

In our work, we use the Refined Schwartz Theory of Basic Human Values \cite{mirzakhmedova2024touche23} as a reference for understanding relationships between different values. Since the theory organizes human values based on how similar or conflicting they are, it provides a natural way to evaluate whether these relationships are reflected in the learned vector representations. See Figure~\ref{fig:schwartz_wheel} for the Schwartz circumplex and Table~\ref{tab:schwartz_values} for the explanation of values.
\cam{We adopt the taxonomy of \citet{kiesel2022identifying}, which extends the refined 19-value framework of \citet{schwartz2012overview} with one additional category, \emph{Universalism: Objectivity}. Based on comparisons with other value inventories, this category is placed between \emph{Universalism: Tolerance} and \emph{Self-Direction: Thought}, thereby preserving the circumplex structure. This 20-category framework has become the standard taxonomy in NLP research on human values~\citep{mirzakhmedova2024touche23,kiesel2023semeval}.}

\begin{table*}[!htbp]
    \centering
    \caption{%
    \textbf{The 20 Basic Human Values of the Refined Schwartz Theory}
    and their core motivational goals.
    }
    \label{tab:schwartz_values}
    \setlength{\tabcolsep}{5pt}
    \renewcommand{\arraystretch}{1.25}
    \small
    \begin{tabular}{l !{\vrule width 0.6pt} l}
        \toprule
        \textbf{Value} & \textbf{Core Motivational Goal} \\
        \midrule
        Self-Direction: Thought
            & Freedom to cultivate one's own ideas and abilities \\
        \cdashline{1-2}[0.8pt/2pt]
        Self-Direction: Action
            & Freedom to determine one's own actions \\
        \cdashline{1-2}[0.8pt/2pt]
        Stimulation
            & Excitement, novelty, and challenge in life \\
        \cdashline{1-2}[0.8pt/2pt]
        Hedonism
            & Pleasure and sensuous gratification for oneself \\
        \cdashline{1-2}[0.8pt/2pt]
        Achievement
            & Success through demonstrating competence by social standards \\
        \cdashline{1-2}[0.8pt/2pt]
        Power: Dominance
            & Power through exercising control over people \\
        \cdashline{1-2}[0.8pt/2pt]
        Power: Resources
            & Power through control of material and social resources \\
        \cdashline{1-2}[0.8pt/2pt]
        Face
            & Security and power through maintaining public image \\
        \cdashline{1-2}[0.8pt/2pt]
        Security: Personal
            & Safety and security in one's immediate environment \\
        \cdashline{1-2}[0.8pt/2pt]
        Security: Societal
            & Safety and stability in the wider society \\
        \cdashline{1-2}[0.8pt/2pt]
        Tradition
            & Maintaining cultural, family, or religious customs \\
        \cdashline{1-2}[0.8pt/2pt]
        Conformity: Rules
            & Compliance with rules, laws, and formal obligations \\
        \cdashline{1-2}[0.8pt/2pt]
        Conformity: Interpersonal
            & Avoidance of upsetting or harming other people \\
        \cdashline{1-2}[0.8pt/2pt]
        Humility
            & Recognizing one's insignificance in the larger scheme \\
        \cdashline{1-2}[0.8pt/2pt]
        Benevolence: Dependability
            & Being a reliable and trustworthy member of the in-group \\
        \cdashline{1-2}[0.8pt/2pt]
        Benevolence: Caring
            & Devotion to the welfare of in-group members \\
        \cdashline{1-2}[0.8pt/2pt]
        Universalism: Concern
            & Commitment to equality and justice for all people \\
        \cdashline{1-2}[0.8pt/2pt]
        Universalism: Nature
            & Preservation of the natural environment \\
        \cdashline{1-2}[0.8pt/2pt]
        Universalism: Tolerance
            & Acceptance and understanding of those who are different \\
        \cdashline{1-2}[0.8pt/2pt]
        Universalism: Objectivity
            & Pursuit of truth, rationality, and an unbiased worldview \\
        \bottomrule
    \end{tabular}
    \label{table:schwartz}
\end{table*}

\section{Additional Results on Llama~3.1-8B}

Tables~\ref{tab:llama31_geometry} and~\ref{tab:llama31_gain} show that the same paradigm-level patterns seen on Qwen~3.5-9B are present on Llama~3.1-8B: distribution-driven techniques outperform behavior-centric methods in recovering the Schwartz circumplex with CAA achieving the strongest theory alignment ($\rho_T = 0.50$, $p < 10^{-12}$). On the other hand, behavior-centric methods have non-trivial behavioral gains but are close to zero or negative for most of the geometry metrics, further highlighting the distinction between behavioral steering performance and geometric fidelity.

\begin{table*}[ht]
    \centering
    \caption{
    \textbf{Effect of steering method on geometry-alignment metrics}
    on the Llama~3.1-8B backbone.
    The metrics include Theory Rank Correlation ($\rho_T$),
    Theory Linear Correlation ($r_T$), Hierarchical Structure Correlation
    ($\rho_H$), and Polarity Separation Score ($\Delta_{\mathrm{pol}}$).
    Subscripts denote p-values. Bold indicates the best value.
    }
    \label{tab:llama31_geometry}
    \setlength{\tabcolsep}{5pt}
    \renewcommand{\arraystretch}{1.25}
    \small
    \resizebox{\textwidth}{!}{\begin{tabular}{l l !{\vrule width 0.6pt} c c c c}
        \toprule
        \textbf{Paradigm} &
        \textbf{Method} &
        $\rho_{\mathrm{T}}$ &
        $r_{\mathrm{T}}$ &
        $\rho_{\mathrm{H}}$ &
        $\Delta_{\mathrm{pol}}$ \\
        \midrule

        \multicolumn{6}{l}{\textit{\textbf{Base} (pre-trained)}} \\[1pt]

        \multirow{7}{*}{\textit{Behav.-centric}}
        & LLM Raw Activation Space
        & $0.2387_{{\scriptscriptstyle(9.1\text{e-}4)}}$
        & $0.2377_{{\scriptscriptstyle(9.6\text{e-}4)}}$
        & $0.2530_{{\scriptscriptstyle(4.3\text{e-}4)}}$
        & 0.0026 \\
        \cdashline{1-6}[0.8pt/2pt]

        & OPT~\cite{dunefsky2025one}
            & $-0.0486_{{\scriptscriptstyle(5.1\text{e-}1)}}$
            & $-0.0696_{{\scriptscriptstyle(3.4\text{e-}1)}}$
            & $-0.0352_{{\scriptscriptstyle(6.3\text{e-}1)}}$
            & 0.0232 \\
        & Cold-Steer (FD)~\cite{sharma2026cold}
            & $-0.0405_{{\scriptscriptstyle(5.8\text{e-}1)}}$
            & $-0.0404_{{\scriptscriptstyle(5.8\text{e-}1)}}$
            & $-0.0365_{{\scriptscriptstyle(6.2\text{e-}1)}}$
            & 0.0104 \\
        & BiPO~\cite{cao2024personalized}
            & $0.1094_{{\scriptscriptstyle(1.3\text{e-}1)}}$
            & $0.1036_{{\scriptscriptstyle(1.5\text{e-}1)}}$
            & $0.0098_{{\scriptscriptstyle(8.9\text{e-}1)}}$
            & $-$0.0094 \\
        \cdashline{1-6}[0.8pt/2pt]

        \multirow{4}{*}{\textit{Dist.-driven}}
        & ODESteer~\cite{zhao2026odesteer}
            & $0.3124_{{\scriptscriptstyle(1.1\text{e-}5)}}$
            & $0.3026_{{\scriptscriptstyle(2.2\text{e-}5)}}$
            & $0.2996_{{\scriptscriptstyle(2.7\text{e-}5)}}$
            & 0.0195 \\
        & SAS~\cite{bayat2025steering}
            & $0.4220_{{\scriptscriptstyle(1.3\text{e-}9)}}$
            & $0.4327_{{\scriptscriptstyle(4.5\text{e-}10)}}$
            & $0.3351_{{\scriptscriptstyle(2.3\text{e-}6)}}$
            & 0.3083 \\
        & SphericalSteer~\cite{you2026spherical}
            & $0.4949_{{\scriptscriptstyle(3.9\text{e-}13)}}$
            & $0.5014_{{\scriptscriptstyle(1.7\text{e-}13)}}$
            & $0.4103_{{\scriptscriptstyle(4.1\text{e-}9)}}$
            & \textbf{0.3974} \\
        & CAA~\cite{rimsky2024steering}
            & $\mathbf{0.4996}_{{\scriptscriptstyle(2.2\text{e-}13)}}$
            & $\mathbf{0.5060}_{{\scriptscriptstyle(9.6\text{e-}14)}}$
            & $\mathbf{0.4205}_{{\scriptscriptstyle(1.5\text{e-}9)}}$
            & 0.3957 \\

        \bottomrule
    \end{tabular}}
\end{table*}

\begin{table}[!htbp]
    \centering
    \caption{%
    \textbf{Effect of steering method on accuracy gain}
    for the Llama~3.1-8B backbone.
    Gain is reported in percentage points (pp) relative to the unsteered baseline.
    Bold indicates the best value.
    }
    \label{tab:llama31_gain}
    \setlength{\tabcolsep}{5pt}
    \renewcommand{\arraystretch}{1.25}
    \small
    \begin{tabular}{l !{\vrule width 0.6pt} c}
        \toprule
        \textbf{Method} & \textbf{Gain (pp)} \\
        \midrule
        OPT~\cite{dunefsky2025one}
            & $+6.03$ \\
        \cdashline{1-2}[0.8pt/2pt]
        BiPO~\cite{cao2024personalized}
            & $+6.30$ \\
        \cdashline{1-2}[0.8pt/2pt]
        ODESteer~\cite{zhao2026odesteer}
            & $+6.60$ \\
        \cdashline{1-2}[0.8pt/2pt]
        Cold-Steer~\cite{sharma2026cold}
            & $+7.50$ \\
        \cdashline{1-2}[0.8pt/2pt]
        SphericalSteer~\cite{you2026spherical}
            & $+17.57$ \\
        \cdashline{1-2}[0.8pt/2pt]
        CAA~\cite{rimsky2024steering}
            & $+17.64$ \\
        \cdashline{1-2}[0.8pt/2pt]
        SAS~\cite{bayat2025steering}
            & $\mathbf{+19.87}$ \\
        \bottomrule
    \end{tabular}
\end{table}

\section{Further Scale Analysis}
\label{app:scale-analysis}

\cam{We extend the scale analysis beyond CAA by evaluating SphericalSteer on Qwen3.5 (0.8B--9B) and Qwen2.5 (7B--32B). As shown in Table~\ref{tab:scale-analysis}, both methods exhibit monotonic improvements in theory alignment across both model families. In comparison, the raw-activation baseline varies only modestly and non-monotonically ($\rho_T=0.16$--$0.22$ on Qwen3.5 and $0.14$--$0.19$ on Qwen2.5), suggesting that the improvement is not merely inherited from richer backbone activations. SAS also improves where pretrained SAEs are available ($\rho_T=0.46$ on Qwen3.5-2B and $0.51$ on Qwen3.5-9B), although broader comparison is limited by SAE availability. Available BiPO results show no reliable scale trend. These results support scale-related improvement for distribution-driven methods rather than for steering methods generally.}

\begin{table}[ht]
\centering
\small
\setlength{\tabcolsep}{5pt}
\resizebox{\columnwidth}{!}{%
\begin{tabular}{llcc}
\toprule
Method & Model & $\rho_T$ & $r_T$ \\
\midrule
CAA~\cite{rimsky2024steering}
    & Qwen3.5-0.8B & 0.2537 & 0.2643 \\
    & Qwen3.5-2B   & 0.3364 & 0.3545 \\
    & Qwen3.5-4B   & 0.4109 & 0.4243 \\
    & Qwen3.5-9B   & 0.4606 & 0.4754 \\
    \cdashline{1-4}[0.8pt/2pt]
    & Qwen2.5-7B   & 0.2370 & 0.2491 \\
    & Qwen2.5-14B  & 0.3133 & 0.3298 \\
    & Qwen2.5-32B  & 0.3347 & 0.3470 \\
\midrule
SphericalSteer~\cite{you2026spherical}
    & Qwen3.5-0.8B & 0.2186 & 0.2296 \\
    & Qwen3.5-2B   & 0.3109 & 0.3306 \\
    & Qwen3.5-4B   & 0.3760 & 0.3899 \\
    & Qwen3.5-9B   & 0.3962 & 0.4061 \\
    \cdashline{1-4}[0.8pt/2pt]
    & Qwen2.5-7B   & 0.2332 & 0.2476 \\
    & Qwen2.5-14B  & 0.2773 & 0.2971 \\
    & Qwen2.5-32B  & 0.3320 & 0.3384 \\
\bottomrule
\end{tabular}%
}
\caption{Effect of model scale on theory-alignment metrics for CAA and SphericalSteer across Qwen3.5 and Qwen2.5. All correlations are statistically significant ($p<3\times10^{-3}$).}
\label{tab:scale-analysis}
\end{table}

\section{Dataset Details}
\label{app:stats}
This section describes the construction, filtering, and distribution of the datasets used to extract and evaluate value vectors across the 20 Schwartz value categories.

\subsection{Gathering and Generation}

\paragraph{ValueBench.}
ValueBench~\citep{ren2024valuebench} is built from 44 established psychological questionnaires and tests spanning personality, social axioms, cognitive systems, and general value theory. Each item comes with a value label and an agreement score to indicate whether the item endorses or opposes that value. 
Since ValueBench covers far more dimensions than the 20 Schwartz values, we used Gemini 3 Flash~\citep{gemini3flash} to map each item to its closest Schwartz value, keeping only items where the mapping was clear. For items missing one answer, we generated the missing answer using Qwen3.5-35B-A3B~\citep{team2026qwen3} with in-context learning, and manually verified a representative subset of the process afterward. After filtering, this gave us 911 samples.

\paragraph{Touché23-ValueEval.}
The Touché23-ValueEval dataset~\citep{mirzakhmedova2024touche23} contains 9,324 arguments from six domains, including religious texts, political discussions, newspaper editorials, and online democracy platforms. Each argument was annotated by three crowdworkers for 54 human values organized into 20 Schwartz categories. Each argument pairs a conclusion with a premise that either supports or opposes it.
We turn each conclusion into a question and use the supporting premise as the positive answer. Negative answers are generated using Gemma-4-26B-A4B-it~\citep{gemma4}, each conditioned on one of six prompt strategies for producing value-neutral responses (see Appendix~\S\ref{sec:appendix:dataset}). We then use the same model in an LLM-as-a-judge framework \cite{gu2024survey} to filter out low-quality samples, leaving around 25.5K final pairs.

\subsection{Dataset Statistics}
Table~\ref{tab:dataset_stats} presents the distribution of contrastive quadruples across the 20 Schwartz value categories and two dataset sources after quality filtering. Touché dominates with 25,517 samples (96.6\%), while ValueBench contributes the remaining 911 (3.4\%), yielding 26,428 quadruples in total.

\begin{table*}[t]
\centering
\small
\caption{Value distribution per Schwartz category across both dataset sources. 
Counts reflect quadruples retained after quality filtering.}
\label{tab:dataset_stats}
\begin{tabular}{lrrr}
\toprule
\textbf{Value category} & \textbf{ValueBench} & \textbf{Touché} & \textbf{Total} \\
\midrule
\multicolumn{4}{l}{\textit{Achievement \& Power}} \\
Achievement                  & 125 & 2,187 & 2,312 \\
Power: dominance             &  37 &   792 &   829 \\
Power: resources             &  10 &   779 &   789 \\
\midrule
\multicolumn{4}{l}{\textit{Benevolence \& Universalism}} \\
Benevolence: caring          &  96 & 2,112 & 2,208 \\
Benevolence: dependability   &  19 & 1,149 & 1,168 \\
Universalism: concern        &  64 & 2,863 & 2,927 \\
Universalism: nature         &  30 &   633 &   663 \\
Universalism: objectivity    &  30 & 1,502 & 1,532 \\
Universalism: tolerance      &  34 &   916 &   950 \\
\midrule
\multicolumn{4}{l}{\textit{Security \& Conformity}} \\
Security: personal           &  68 & 2,950 & 3,018 \\
Security: societal           &  10 & 2,277 & 2,287 \\
Conformity: interpersonal    &  37 &   299 &   336 \\
Conformity: rules            &  57 & 1,692 & 1,749 \\
\midrule
\multicolumn{4}{l}{\textit{Self-direction, Stimulation \& Hedonism}} \\
Self-direction: action       &  62 & 1,793 & 1,855 \\
Self-direction: thought      &  90 & 1,087 & 1,177 \\
Stimulation                  &  28 &   418 &   446 \\
Hedonism                     &  11 &   285 &   296 \\
\midrule
\multicolumn{4}{l}{\textit{Tradition, Humility \& Face}} \\
Tradition                    &  52 &   706 &   758 \\
Humility                     &  31 &   550 &   581 \\
Face                         &  20 &   527 &   547 \\
\midrule
\textbf{Total}               & \textbf{911} & \textbf{25,517} & \textbf{26,428} \\
\bottomrule
\end{tabular}
\end{table*}

\subsection{Negative Answer Generation Strategies}
\label{sec:appendix:dataset}
Negative answers for the Touché-derived subset are generated using one of six strategies: pragmatic, which challenges feasibility, cost, or practical implementation; empirical, which contests factual or causal claims using evidence or known outcomes; counter-example, which cites a real or plausible case where the same policy led to opposite results; side-effects, which argues that the policy produces serious unintended consequences in a different domain; institutional, which asserts that existing rules or institutions already address the concern more effectively; and contradict, which disputes whether the value-based premise is valid, relevant, or accurately applied in context.

Each example was deterministically assigned one of the six strategies based on its argument ID and target value; the complete prompts, including all templates, demonstrations, and leakage warnings, are provided in the released code.

\begin{figure*}[!htbp]
    \centering
    \includegraphics[width=1\textwidth]{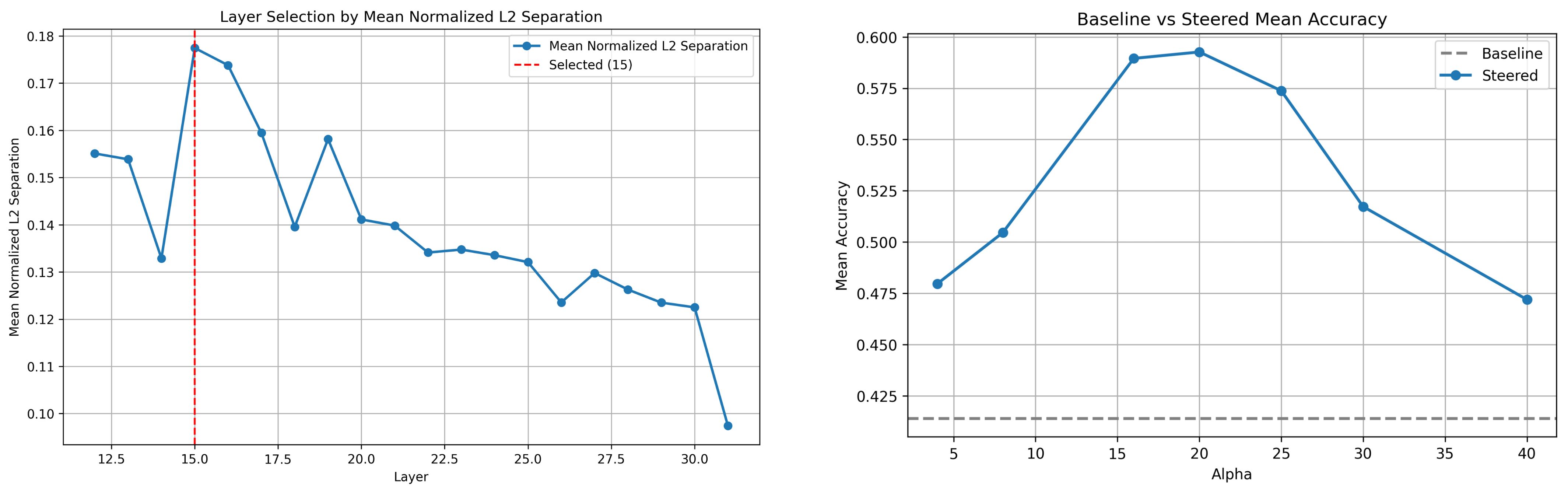}
    \caption{(Left) Layer selection via Mean Normalized L2 Separation across transformer layers for CAA for Qwen3.5-9B-Base. Layer 15 is selected (dashed red line) as it achieves the highest mean separation between positive and negative activation distributions. (Right) Mean steering accuracy across Schwartz values as a function of the multiplier $\alpha$, compared against the unsteered baseline (dashed gray line). Accuracy peaks around $\alpha$ $\in$ [15, 20] before degrading at larger values, consistent with prior observations that excessive steering strength degrades generation quality.}
    \label{fig:accuracy_validation}
\end{figure*}

\section{Robustness of Value-Geometry Evaluation}

\cam{We conduct additional analyses to test whether the observed value geometry and the separation between steering paradigms could be explained by properties of the dataset rather than the extracted value directions. Specifically, we examine robustness to sample selection, surface wording, and multi-value annotations. For each robustness setting, we re-extract steering vectors for all seven methods on Qwen3.5-9B-Base and Llama3.1-8B and repeat the geometry analysis.}

\label{app:robustness_surface}
\subsection{Sample Selection and Surface Wording}

\cam{Most examples in our benchmark are derived from Touch\'e23-ValueEval,
where the question and value-aligned answer originate from the human-annotated
source data and the value-neutral answer is generated by an LLM. We therefore
consider several controls for potential artifacts introduced by this
construction procedure.}

\cam{First, generation is constrained to produce answers of approximately the
same length as their human-written counterparts. Generated answers contain
26.8 tokens on average, compared with 22.2 tokens for the corresponding
human-written answers. We also generate negative answers using six different
prompting strategies---pragmatic, empirical, counter-example, side-effects,
institutional, and contradict---to avoid reliance on a single stylistic
template (Appendix~\S\ref{sec:appendix:dataset}). In addition, different
models are used across stages of the data-construction pipeline, and the
generation models differ from the backbones evaluated in our steering
experiments.}

\cam{We additionally perform a manual audit of 100 generated examples,
sampling 5 examples from each Schwartz value. Overall, 89\% of the generated
answers were judged valid. The primary failure modes were value leakage and
insufficient contrast with the corresponding human-written answer.}

\paragraph{Disjoint resampling.}
\cam{To test whether the observed geometry depends on the particular examples
used in the main experiments, we construct a fresh subset containing 200
examples per value with no overlap with the original subset. We then
re-extract all steering vectors and repeat the complete geometry analysis for
both backbones.}

\paragraph{Surface-form perturbation.}
\cam{To test sensitivity to lexical and stylistic cues, we construct a second
variant in which the question and both contrastive answers of every example
are paraphrased using \texttt{gemma-4-26B-A4B-it}. The paraphrasing procedure
preserves the semantic content and target-value assignment while changing
the surface realization of all three components. We again re-extract the
steering vectors from the paraphrased data and repeat the full geometry
evaluation.}

\cam{Across both the disjoint and paraphrased conditions, the main
paradigm-level separation is preserved: distribution-driven methods retain
substantial theory alignment, whereas behavior-centric methods remain close
to zero. In particular, CAA and SAS remain statistically significant across
both backbones and all three data settings. These results indicate that the
observed separation is not specific to the original sample selection or
surface realization of the contrastive examples.}

\subsection{Multi-Value Annotation Sensitivity}
\label{app:robustness_multivalue}

\cam{Real-world arguments may express several human values simultaneously,
and the source annotations can therefore assign multiple values to the same
example. Consequently, we interpret the extracted direction for a target
value as a \emph{target-conditioned aggregate direction}, rather than as a
perfectly monosemantic representation of that value.}

\cam{To examine whether multi-value annotations drive the measured geometry,
we construct an additional subset that prioritizes examples annotated with
exactly one value in the source data. We select 200 examples per Schwartz
value, using single-label examples whenever available. Because some values
do not contain 200 such examples, shortages are filled with different
multi-label examples that were not used in the original subset. The resulting
dataset is approximately 50\% single-label overall. Since the availability
of single-label examples differs across values and this procedure also
changes the sample distribution, we treat this experiment as a conservative
sensitivity analysis rather than a pure single-label ablation.}

\cam{The paradigm-level separation also remains under the
single-label-prioritized sensitivity analysis on both backbones
(Table~\ref{tab:robustness-all}). On Qwen3.5-9B-Base, CAA, SAS,
SphericalSteer, and ODESteer obtain $\rho_T$ values of 0.325, 0.309,
0.355, and 0.256, respectively, compared with 0.122 for the matched
raw-activation baseline and 0.010, 0.074, and 0.055 for BiPO, OPT, and
COLD-Steer. On Llama3.1-8B, the corresponding distribution-driven methods
obtain $\rho_T$ values of 0.430, 0.337, 0.442, and 0.262, compared with
0.126 for the raw-activation baseline and 0.054, 0.033, and 0.062 for
BiPO, OPT, and COLD-Steer.}

\cam{Absolute correlations are lower than in the main experiment for several
methods, but the matched raw-activation baseline also decreases under this
sampling procedure. Thus, the reduction is not specific to the steering
methods and is consistent with the change in sample composition. Most
importantly, the separation between distribution-driven and
behavior-centric methods and the associated statistical significance are
preserved on both backbones. This sensitivity analysis therefore supports
the robustness of our paradigm-level conclusion to multi-value annotation,
while not implying that the extracted directions are perfectly
monosemantic.}

\begin{table*}[!htbp]
\centering
\caption{
Robustness of Theory Rank Correlation ($\rho_{\mathrm{T}}$) to dataset
construction. \textit{Original} denotes the main experimental setting;
\textit{Disjoint} uses a fresh non-overlapping subset of 200 examples per
value; \textit{Paraphrased} paraphrases the question and both contrastive
answers; and \textit{SL-prio.} uses the
single-label-prioritized sensitivity subset. Subscripts denote $p$-values.
The paradigm-level separation is preserved across all robustness settings.
Bold indicates the highest correlation within each setting.
}
\label{tab:robustness-all}
\setlength{\tabcolsep}{4pt}
\renewcommand{\arraystretch}{1.25}
\small
\resizebox{\textwidth}{!}{
\begin{tabular}{lcccc|cccc}
\toprule
& \multicolumn{4}{c|}{\textbf{Qwen3.5-9B-Base}}
& \multicolumn{4}{c}{\textbf{Llama~3.1-8B}} \\
\cmidrule(lr){2-5}
\cmidrule(lr){6-9}
\textbf{Method}
& \textbf{Original}
& \textbf{Disjoint}
& \textbf{Paraphrased}
& \textbf{SL-prio.}
& \textbf{Original}
& \textbf{Disjoint}
& \textbf{Paraphrased}
& \textbf{SL-prio.} \\
\midrule

LLM Raw Activation Space
& $0.2228_{{\scriptscriptstyle(2.0\text{e-}3)}}$
& $0.2122_{{\scriptscriptstyle(3.3\text{e-}3)}}$
& $0.2362_{{\scriptscriptstyle(1.0\text{e-}3)}}$
& $0.1223_{{\scriptscriptstyle(9.3\text{e-}2)}}$
& $0.2387_{{\scriptscriptstyle(9.1\text{e-}4)}}$
& $0.2400_{{\scriptscriptstyle(8.5\text{e-}4)}}$
& $0.2305_{{\scriptscriptstyle(1.4\text{e-}3)}}$
& $0.1260_{{\scriptscriptstyle(8.3\text{e-}2)}}$ \\

\midrule
\multicolumn{9}{l}{\textit{Behavior-centric}} \\

OPT~\cite{dunefsky2025one}
& $0.1138_{{\scriptscriptstyle(1.2\text{e-}1)}}$
& $0.0824_{{\scriptscriptstyle(2.6\text{e-}1)}}$
& $-0.0416_{{\scriptscriptstyle(5.7\text{e-}1)}}$
& $0.0740_{{\scriptscriptstyle(3.1\text{e-}1)}}$
& $-0.0486_{{\scriptscriptstyle(5.1\text{e-}1)}}$
& $0.0961_{{\scriptscriptstyle(1.9\text{e-}1)}}$
& $-0.0294_{{\scriptscriptstyle(6.9\text{e-}1)}}$
& $0.0335_{{\scriptscriptstyle(6.5\text{e-}1)}}$ \\

COLD-Steer (FD)~\cite{sharma2026cold}
& $0.0309_{{\scriptscriptstyle(9.1\text{e-}1)}}$
& $-0.0877_{{\scriptscriptstyle(2.3\text{e-}1)}}$
& $-0.0622_{{\scriptscriptstyle(3.9\text{e-}1)}}$
& $0.0553_{{\scriptscriptstyle(4.5\text{e-}1)}}$
& $-0.0405_{{\scriptscriptstyle(5.8\text{e-}1)}}$
& $0.0820_{{\scriptscriptstyle(2.6\text{e-}1)}}$
& $-0.0122_{{\scriptscriptstyle(8.7\text{e-}1)}}$
& $0.0615_{{\scriptscriptstyle(4.0\text{e-}1)}}$ \\

BiPO~\cite{cao2024personalized}
& $0.1188_{{\scriptscriptstyle(1.0\text{e-}1)}}$
& $0.0491_{{\scriptscriptstyle(5.0\text{e-}1)}}$
& $-0.0666_{{\scriptscriptstyle(3.6\text{e-}1)}}$
& $0.0099_{{\scriptscriptstyle(8.9\text{e-}1)}}$
& $0.1094_{{\scriptscriptstyle(1.3\text{e-}1)}}$
& $0.0532_{{\scriptscriptstyle(4.7\text{e-}1)}}$
& $0.0279_{{\scriptscriptstyle(7.0\text{e-}1)}}$
& $0.0539_{{\scriptscriptstyle(4.6\text{e-}1)}}$ \\

\midrule
\multicolumn{9}{l}{\textit{Distribution-driven}} \\

ODESteer~\cite{zhao2026odesteer}
& $0.2741_{{\scriptscriptstyle(1.3\text{e-}4)}}$
& $0.3029_{{\scriptscriptstyle(2.2\text{e-}5)}}$
& $0.4325_{{\scriptscriptstyle(4.6\text{e-}10)}}$
& $0.2560_{{\scriptscriptstyle(3.6\text{e-}4)}}$
& $0.3124_{{\scriptscriptstyle(1.1\text{e-}5)}}$
& $0.2094_{{\scriptscriptstyle(3.7\text{e-}3)}}$
& $0.2881_{{\scriptscriptstyle(5.5\text{e-}5)}}$
& $0.2624_{{\scriptscriptstyle(2.6\text{e-}4)}}$ \\

SphericalSteer~\cite{you2026spherical}
& $0.3962_{{\scriptscriptstyle(1.5\text{e-}8)}}$
& $0.4112_{{\scriptscriptstyle(3.8\text{e-}9)}}$
& $0.4265_{{\scriptscriptstyle(8.5\text{e-}10)}}$
& $\mathbf{0.3554}_{{\scriptscriptstyle(4.9\text{e-}7)}}$
& $0.4949_{{\scriptscriptstyle(3.9\text{e-}13)}}$
& $0.5264_{{\scriptscriptstyle(6.2\text{e-}15)}}$
& $\mathbf{0.4955}_{{\scriptscriptstyle(3.6\text{e-}13)}}$
& $\mathbf{0.4416}_{{\scriptscriptstyle(1.8\text{e-}10)}}$ \\

CAA~\cite{rimsky2024steering}
& $0.4606_{{\scriptscriptstyle(2.3\text{e-}11)}}$
& $0.4365_{{\scriptscriptstyle(3.1\text{e-}10)}}$
& $\mathbf{0.4634}_{{\scriptscriptstyle(1.7\text{e-}11)}}$
& $0.3249_{{\scriptscriptstyle(4.8\text{e-}6)}}$
& $\mathbf{0.4996}_{{\scriptscriptstyle(2.2\text{e-}13)}}$
& $\mathbf{0.5277}_{{\scriptscriptstyle(5.2\text{e-}15)}}$
& $0.4902_{{\scriptscriptstyle(7.0\text{e-}13)}}$
& $0.4304_{{\scriptscriptstyle(5.7\text{e-}10)}}$ \\

SAS~\cite{bayat2025steering}
& $\mathbf{0.5069}_{{\scriptscriptstyle(8.5\text{e-}14)}}$
& $\mathbf{0.4514}_{{\scriptscriptstyle(6.3\text{e-}11)}}$
& $0.4381_{{\scriptscriptstyle(2.6\text{e-}10)}}$
& $0.3090_{{\scriptscriptstyle(1.4\text{e-}5)}}$
& $0.4220_{{\scriptscriptstyle(1.3\text{e-}9)}}$
& $0.4515_{{\scriptscriptstyle(6.3\text{e-}11)}}$
& $0.3918_{{\scriptscriptstyle(2.3\text{e-}8)}}$
& $0.3370_{{\scriptscriptstyle(2.0\text{e-}6)}}$ \\

\bottomrule
\end{tabular}
}
\end{table*}

\section{Moral Foundations Theory Evaluation}
\label{app:mft}
We chose Schwartz's theory as the primary framework because its cross-culturally validated circumplex specifies explicit pairwise compatibility and opposition in continuous and discrete form, giving a human-derived geometric ground truth for detailed analysis. Other frameworks do not offer this: Moral Foundations Theory (MFT), as the most prominent alternative, defines six discrete foundations but doesn’t specify the relation between them; its only theory-specified structure is the Individualizing–Binding split.
\subsection{Benchmark Construction}

\cam{We adopt the six-foundation revised Moral Foundations Theory (MFT) of \citet{atari-etal-2023-morality}: Care, Equality, Proportionality, Loyalty, Authority, and Purity. We source the initial samples from the Moral Foundations Reddit Corpus \citep{trager-etal-2026-moral}. We retain only high-confidence, single-foundation samples with full agreement on the foundation label among at least three trained annotators. Using the same generation and filtering pipeline as our Schwartz benchmark, we convert the retained Reddit samples into question-contrastive instances. We balance the benchmark at 200 samples per foundation, yielding 1,200 samples: 400 from the Individualizing foundations and 800 from the Binding foundations.}

\subsection{Family-Structure Metric}
\cam{Unlike Schwartz's theory, MFT specifies neither a circumplex nor inter-foundation angles or opposing pairs; therefore, no analogue of the Schwartz theoretical similarity matrix is available. We instead evaluate MFT's theory-specified higher-order grouping into Individualizing foundations $\{\text{Care}, \text{Equality}\}$ and Binding foundations $\{\text{Proportionality}, \text{Loyalty}, \text{Authority}, \text{Purity}\}$.}

\cam{Let $\mathcal{W}$ denote the seven unordered within-family pairs (one Individualizing pair and six Binding pairs) and let $\mathcal{C}$ denote the eight cross-family pairs. For unit-normalized foundation vectors $u_i$, we define their empirical cosine similarity as $E_{ij}=u_i^\top u_j$. Analogous to $\Delta_{\mathrm{pol}}$, MFT family separation is}
\begin{equation}
\Delta_{\mathrm{MFT}}
=
\frac{1}{|\mathcal{W}|}
\sum_{(i,j)\in\mathcal{W}} E_{ij}
-
\frac{1}{|\mathcal{C}|}
\sum_{(i,j)\in\mathcal{C}} E_{ij}.
\end{equation}
\cam{Positive $\Delta_{\mathrm{MFT}}$ indicates that foundations within the same MFT family are more similar than foundations across families. As with $\Delta_{\mathrm{pol}}$, we report $\Delta_{\mathrm{MFT}}$ as an interpretable separation score without an associated $p$-value.}

\section{Geometry Metrics}
\label{sec:appendix_metrics}

\paragraph{Evaluation Metric Overview.}
\label{app:metric-overview}
\cam{Because validating value geometry has no single established metric, we use complementary measures covering both the geometry of the extracted vectors and its behavioral consequences. The metrics are organized along two dimensions: value-vector geometry versus cross-value transfer, and continuous circumplex structure versus discrete hierarchical structure (Summary in Table \ref{tab:metric-summary}).}

\begin{table*}[t]
\centering
\small
\setlength{\tabcolsep}{3.5pt}
\renewcommand{\arraystretch}{1.16}
\begin{tabular}{
    >{\raggedright\arraybackslash}p{0.21\textwidth}
    >{\raggedright\arraybackslash}p{0.15\textwidth}
    >{\raggedright\arraybackslash}p{0.45\textwidth}
    >{\raggedright\arraybackslash}p{0.10\textwidth}
}
\toprule
\textbf{Metric}
& \textbf{Structure tested}
& \textbf{What it measures}
& \textbf{Defined in} \\
\midrule

\rowcolor{gray!12}
\multicolumn{4}{l}{\textbf{Geometry of value vectors}} \\

Theory Rank Correlation ($\rho_T$)
& Continuous circumplex
& Whether compatible values rank above conflicting values in vector similarity
& \makecell[l]{Sec.~\ref{sec:geometry_eval}\\App.~\S\ref{sec:appendix_metrics}} \\

Theory Linear Correlation ($r_T$)
& Continuous circumplex
& Whether empirical cosine similarities scale linearly with theoretical circumplex similarities
& \makecell[l]{Sec.~\ref{sec:geometry_eval}\\App.~\S\ref{sec:appendix_metrics}} \\

Hierarchical Structure Correlation ($\rho_H$)
& Discrete hierarchy
& Whether vector similarities respect lower-order families and higher-order groups
& \makecell[l]{Sec.~\ref{sec:geometry_eval}\\App.~\S\ref{sec:appendix_metrics}} \\

Polarity Separation Score ($\Delta_{\mathrm{pol}}$)
& Same-family vs.\ opposing groups
& The difference between mean within-family similarity and mean similarity across opposing groups
& \makecell[l]{Sec.~\ref{sec:geometry_eval}\\App.~\S\ref{sec:appendix_metrics}} \\

\midrule
\rowcolor{gray!12}
\multicolumn{4}{l}{\textbf{Cross-value transfer}} \\

Continuous Transfer Fidelity (TWTM)
& Continuous circumplex
& Whether transfer is positive for neighboring values and negative for opposing values, accounting for both direction and magnitude
& Sec.~\ref{sec:transfer_eval} \\

Hierarchical Transfer Fidelity ($\rho_H^{\mathrm{tr}}$)
& Discrete hierarchy
& Whether transfer ranks highest within value families and lowest across opposing groups
& Sec.~\ref{sec:transfer_eval} \\

Bin-wise Monotonic Decay (BMD-$\rho$)
& Continuous circumplex
& Whether mean transfer decreases monotonically with circumplex distance, independent of magnitude
& App.~\S\ref{app:bmd} \\

\bottomrule
\end{tabular}
\caption{Summary of the evaluation metrics and the theoretical structure assessed by each metric. Geometry metrics evaluate relationships among extracted value vectors, while transfer metrics evaluate how steering effects propagate across values.}
\label{tab:metric-summary}
\end{table*}

Let $\mathbf{u}_1, \ldots, \mathbf{u}_{20}$ be unit-normalized steering vectors for the 20 Schwartz values, indexed in their canonical circumplex order. The \textbf{empirical similarity matrix} is $E_{bb'} = \mathbf{u}_b^\top \mathbf{u}_{b'}$, and $E_\triangle$ denotes its 190 upper-triangle entries. 

\paragraph{Construction of the Theoretical Similarity Matrix.}
\label{app:theoretical-matrix}

\cam{The theoretical matrix follows directly from the central assumption of Schwartz's theory: values occupy a fixed canonical order on a circumplex, and the compatibility between two values is determined by their angular distance on this circle~\citep{schwartz1992universals,schwartz2012overview}. With 20 values, adjacent positions are separated by $18^\circ$, and $T_{bb'}$ is defined as the cosine of the shortest angular distance between values $b$ and $b'$:
\[
T_{bb'} =
\cos\!\left(
\min(|b-b'|,\,20-|b-b'|)\times18^\circ
\right).
\]
Consequently, adjacent values, such as \emph{Benevolence: Caring} and \emph{Universalism: Concern}, receive a theoretical similarity close to $1$, whereas diametrically opposed values receive $-1$. This cosine-of-angular-distance construction operationalizes the circumplex structure established in the psychology literature through cross-cultural multidimensional-scaling analyses. For the Hierarchical Structure Correlation and Polarity Separation Score, we additionally use the circumplex's higher-order groups (Openness to Change, Self-Enhancement, Conservation, and Self-Transcendence) shown in Figure~\ref{fig:schwartz_wheel}.}
\cam{Since the Schwartz circumplex is an idealized theoretical model that is itself recovered only approximately from human questionnaire data through smallest-space analysis, perfect correspondence with a cosine-similarity matrix is not expected~\citep{schwartz1992universals}.}

\paragraph{Theory Rank Correlation ($\rho_{\mathrm{T}}$).}
\[
  \rho_{\mathrm{T}} = \mathrm{Spearman}(E_\triangle,\; T_\triangle).
\]
\paragraph{Theory Linear Correlation ($r_{\mathrm{T}}$).}
\[
  r_{\mathrm{T}} = \mathrm{Pearson}(E_\triangle,\; T_\triangle).
\]
Both $\rho_{\mathrm{T}}$ and $r_{\mathrm{T}}$ are reported with p-values from standard two-tailed tests over $n=190$ observations.
\paragraph{Hierarchical Structure Correlation ($\rho_{\mathrm{H}}$).}
The theory-based hierarchical distance is:
\[
d^{\mathrm{hier}}_{bb'} =
\resizebox{0.88\columnwidth}{!}{$
\begin{cases}
    1  & \text{same lower-order family,}\\
    2  & \text{same higher-order group, different family,}\\
    5  & \text{no clear relation,}\\
    10 & \text{opposing higher-order groups.}
\end{cases}
$}
\]
Boundary values (Hedonism, Face, Humility) belong to two higher-order groups simultaneously. Then:
\[
  \rho_{\mathrm{H}} = \mathrm{Spearman}(E_\triangle,\; -d^{\mathrm{hier}}_\triangle).
\]
$\rho_{\mathrm{H}}$ is reported with a p-value.
\paragraph{Polarity Separation Score ($\Delta_{\mathrm{pol}}$).}
Let $\mathcal{S}^{-}$ be the set of same lower-order family pairs and $\mathcal{S}^{\oplus}$ the set of opposing higher-order group pairs. Then:
\[
  \Delta_{\mathrm{pol}}
  = \frac{1}{|\mathcal{S}^{-}|}\sum_{(b,b')\in\mathcal{S}^{-}} E_{bb'}
  \;-\;
  \frac{1}{|\mathcal{S}^{\oplus}|}\sum_{(b,b')\in\mathcal{S}^{\oplus}} E_{bb'}.
\]
No p-value is reported for $\Delta_{\mathrm{pol}}$; it serves as an interpretable complement to the correlation metrics.

\section{Bin-wise Monotonic Decay (BMD-\texorpdfstring{$\rho$}{rho})}
\label{app:bmd}
We report a complementary measure of cross-value transfer that disregards magnitude and focuses purely on whether transfer decays monotonically with circumplex distance.

\paragraph{Definition.}
We partition the 380 off-diagonal entries of $T_{\mathrm{res}}$ into ten bins by their 
circumplex step distance $k \in \{1, \ldots, 10\}$, where $k=1$ denotes 
adjacent values and $k=10$ denotes 
diametrically opposing values. For each bin, we 
compute the mean residual transfer:
\[
    \mu_k = \frac{1}{|\mathcal{P}_k|}\sum_{(A,B)\,\in\,\mathcal{P}_k} 
    T_{\mathrm{res}}[A, B],
\]
where $\mathcal{P}_k$ is the set of off-diagonal pairs at distance $k$ 
(40 pairs for $k = 1,\ldots,9$; 20 pairs for $k = 10$). The Schwartz 
circumplex predicts that $\mu_k$ should decrease monotonically with $k$: 
steering toward value $A$ should on average lift values nearby on the 
circumplex ($\mu_1 > 0$) and suppress values on the opposite side 
($\mu_{10} < 0$), with a smooth decay in between.
To put this prediction into practice, \textbf{BMD-$\rho$} measures the negative Spearman correlation between $k$ and $(\mu_1, \ldots, \mu_{10})$:
\begin{equation}
\mathrm{BMD\text{-}\rho} = -\mathrm{Spearman}\big(k,\; \mu_k\big).
\end{equation}
Higher values reflect the theoretically predicted decay from positive transfer at adjacent pairs ($k=1$) to negative transfer at opposing pairs ($k=10$). Unlike TWTM, BMD-$\rho$ is insensitive to absolute magnitude since a method producing tiny but correctly-ordered transfers can score as high as one producing large effects, provided the ordering across bins respects the circumplex.

\begin{figure}[h]
    \centering
    \includegraphics[width=\columnwidth]{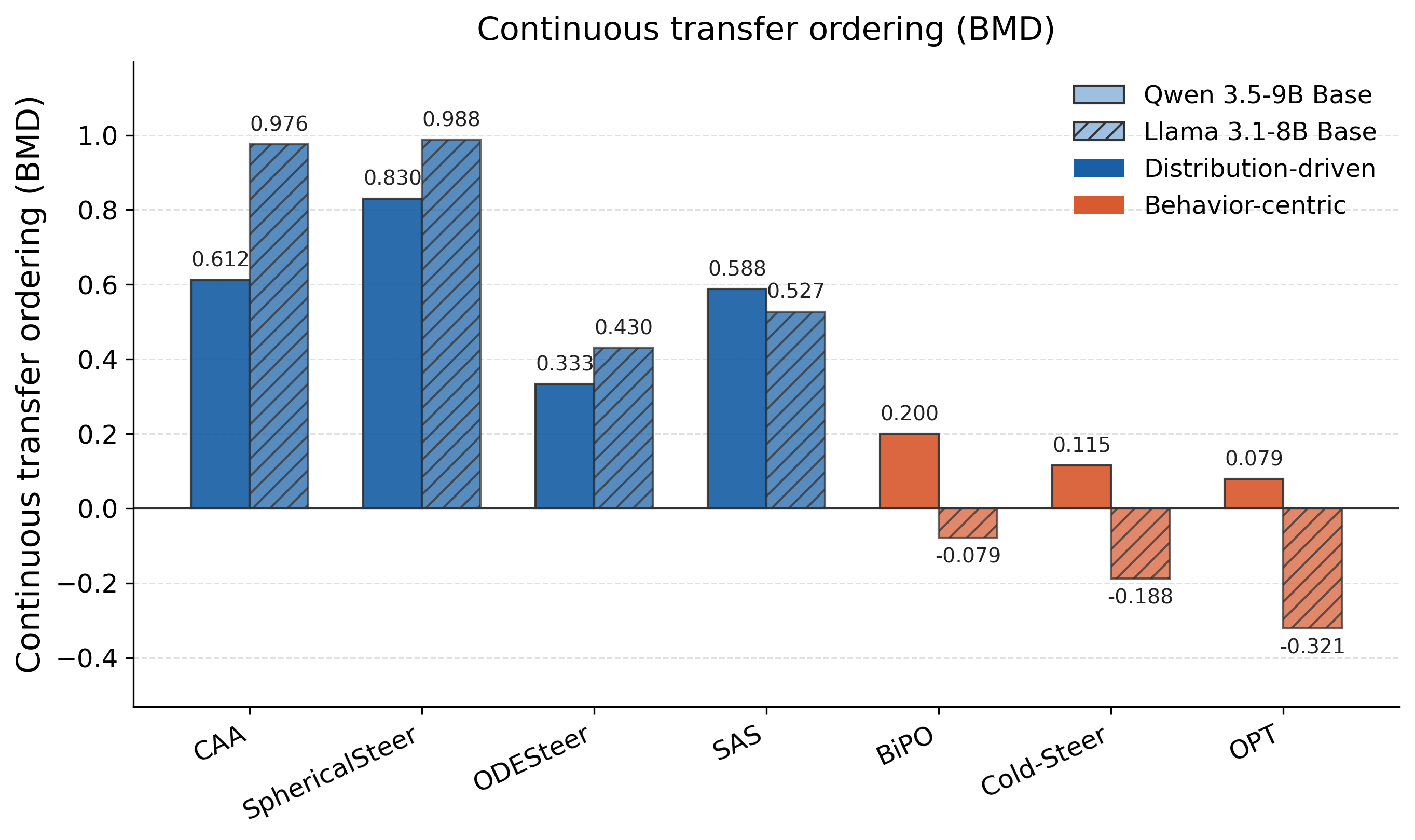}
    \caption{Residualized BMD-$\rho$ across steering methods on Qwen3.5-9B-Base and Llama3.1-8B, grouped by paradigm. Distribution-driven methods (blue) produce transfer patterns whose ordering decays monotonically with circumplex distance, while behavior-centric methods (orange) do not.}
    \label{fig:bmd_bars}
\end{figure}

\paragraph{Results.}
Figure~\ref{fig:bmd_bars} reports BMD-$\rho$ for all seven methods on 
both Qwen3.5-9B-Base and Llama3.1-8B. Two findings stand out.
First, the paradigm separation from the main TWTM result is reproduced and strengthened here. On Qwen, all four distribution-driven methods score between 0.33 and 0.83, while all three behavior-centric methods score between 0.08 and 0.20, which represents a non-overlapping separation. The lowest distribution-driven method (ODESteer, 0.33) exceeds the highest behavior-centric method (BiPO, 0.20) by a clear margin. OPT (0.08) and Cold-Steer (0.12) approach near zero, meaning their bin-wise transfer ordering does not support a monotonic relationship to circumplex distance. This confirms that the paradigm split observed in TWTM is not an artifact of magnitude differences but reflects a genuine structural difference in how the two families of methods propagate steering effects across the value space.

Second, the Llama results reveal that this structural separation is 
not model-specific. Distribution-driven methods on Llama score 0.43--0.99, while Behavior-centric methods collapse to negative BMD-$\rho$, meaning their transfer ordering is actively 
inverted relative to the circumplex on this model. A method with 
negative BMD-$\rho$ does not merely fail to respect Schwartz structure; rather, it systematically transfers more to opposing values than to adjacent ones, which is the opposite of what human value theory predicts.

\section{Steering Method Formulations}
\label{sec:appendix_implementation}
The descriptions below detail the formulations used to extract and apply the value vector $v_{(b,\ell)}$ for each baseline method. 

\paragraph{Intuition Behind the Two Steering Paradigms.}
\cam{The key difference in these two paradigms is what supervises the vector. Distribution-driven methods derive it from where activations of contrastive pairs lie in activation space (e.g., CAA averages their difference), so the vector inherits the model's internal representation of the value. Behavior-centric methods instead optimize the vector against an output objective (e.g., OPT maximizes the log-probability of the aligned answer), so any direction that flips the output is a valid solution, whether or not it corresponds to how the model represents the value. This is why the two classes can achieve similar target accuracy yet show very different geometric properties.}

\paragraph{Unified Effective-Shift Representation.}
\label{app:effective-shift}
\cam{The effective shift in Equation~\ref{eq:effective-shift} 
is defined as the difference between the steered and unsteered activations, so it measures exactly the intervention each method applies, independent of internal mechanism. We choose this because methods such as ODESteer (a nonlinear ODE trajectory) and SphericalSteer (a norm-preserving rotation) do not expose a native steering vector, and this was the only representation that treats every method on equal footing.
For OPT and BiPO, the intervention is a fixed additive vector, so the effective shift recovers it exactly; for COLD-Steer, following its original formulation, we average the finite-difference displacement over the training prompts. Thus, this representation cannot systematically favor distribution-driven methods: if anything, collapsing an intervention into a single vector may disadvantage the non-additive ODESteer and SphericalSteer methods by discarding information about their trajectories and rotations, yet both retain clear theory-aligned structure. Independently, our cross-value transfer evaluation is behavioral: it measures accuracy changes on held-out values without using the pairwise geometry of the effective-shift vectors. The same paradigm split appears there, which further confirms the findings. 
}

\paragraph{LLM Raw Activation Space.} This non-steering baseline measures the geometry already present in the unmodified model. For each Schwartz value, we format the multiple-choice prompt so that the value-aligned option is the target answer, run the backbone without interventions, and extract the last-token hidden activation at the selected layer. The representation for each value is the mean of these positive-answer activations over the corresponding examples. We then compute the same pairwise cosine-similarity geometry metrics used for value vectors, allowing us to compare learned or constructed steering directions against the model's native value representation.

\paragraph{Contrastive Activation Addition (CAA).} Following \cite{rimsky2024steering}, given a contrastive dataset $D_b = \{(p_i, c_i^+, c_i^-)\}$ for each Schwartz value $b$, and a given layer $\ell$, the value vector is defined as the mean difference in residual stream activations:
\begin{equation}
v_{(b, \ell)} = \frac{1}{|D_b|} \sum_{(p_i, c_i^+, c_i^-) \in D_b} [a_\ell(p_i, c_i^+) - a_\ell(p_i, c_i^-)]
\end{equation}
At inference time, the vector is unit-normalized and injected into the residual stream at the token position where the LLM generates the answer as $a_\ell \leftarrow a_\ell + \alpha v_{(b, \ell)}$, where $\alpha$ controls steering strength. 

\paragraph{Optimization Steering (OPT).} As an alternative to contrastive averaging, we directly learn value vectors via gradient descent, following \cite{dunefsky2025one}. Concretely, we employ mixed steering, minimizing the combined loss:
\begin{equation}
\resizebox{1\columnwidth}{!}{%
$\mathcal{L}(v) = \underbrace{-\sum_k \log P_v(c_k^+ \mid p)}_{\text{promotion}} + \underbrace{-\sum_k \log (1 - P_v(c_k^- \mid p))}_{\text{suppression}}$%
}
\end{equation}
where $P_v$ denotes the model's output distribution under steered activations. This formulation simultaneously promotes value-aligned responses and suppresses value-contrary ones. The saved optimized vector is then applied additively at the selected layer, preserving its learned norm: $a_\ell \leftarrow a_\ell + \alpha v_{(b, \ell)}$.

\paragraph{COLD-Steer (Finite-Difference).} For the COLD-Steer finite-difference baseline \cite{sharma2026cold}, each value first induces a small parameter-space perturbation from the gradient of the value-aligned training loss. If $\theta'=\theta+\epsilon\nabla_\theta \mathcal{L}$, the native COLD intervention estimates an activation-space finite-difference direction $(z_{\theta'}(x)-z_\theta(x))/\epsilon$ and steers by subtracting a scaled version of this direction. For geometry comparison, our code stores a representative per-value vector by averaging this finite-difference displacement at the last response token over the training prompts; evaluation applies the saved direction additively at the selected layer with the sign and scale matching the native COLD hook.

\paragraph{Bi-directional Preference Optimization (BiPO).} BiPO \cite{cao2024personalized} learns a bidirectional activation vector from preference pairs. The reference implementation wraps a selected transformer block with a trainable vector addition, freezes the base model, and optimizes only this vector with a DPO-style chosen/rejected objective. During training, the vector multiplier is sampled from $\{+1,-1\}$, and the loss sign is flipped for negative multipliers, encouraging a single direction whose positive and negative scales steer toward opposite preferences. In our comparison adapter, the resulting saved vector is applied with the same additive layer intervention, $a_\ell \leftarrow a_\ell + \alpha v_{(b, \ell)}$, without unit-normalizing the learned magnitude.

\paragraph{ODESteer.} ODESteer \cite{zhao2026odesteer} treats steering as motion through activation space rather than a fixed additive offset. For each value, it fits a polynomial sketch classifier separating value-aligned and value-contrary last-token activations. The normalized gradient of this classifier defines a vector field $f(x)$ pointing toward the aligned region, and steering transforms an activation by integrating the ordinary differential equation $dx/dt=f(x)$ for time $T=\alpha$ using a fixed-step solver. In our exact evaluation mode, one ODESteer model is fitted per Schwartz value and hooked at the selected layer; the steered hidden state is the nonlinear ODE-transformed activation rather than $x+\alpha v$.

\paragraph{SphericalSteer.} SphericalSteer \cite{you2026spherical} uses the CAA mean-difference direction as a unit prototype on the activation sphere. For each value, $\mu_T$ is the normalized positive-minus-negative prototype and $-\mu_T$ is treated as the antipodal contrary prototype. At inference time, the method compares the activation's von Mises-Fisher-style affinity to the two prototypes; when the contrary pole is sufficiently favored, it rotates the activation along the sphere toward $\mu_T$ by a strength controlled by $\alpha$, while preserving the original activation norm. Thus, SphericalSteer modifies angular position rather than adding a Euclidean offset.

\paragraph{Sparse Activation Steering (SAS).} We adapt the SAS framework \cite{bayat2025steering} to operate within the Qwen-Scope TopK SAE feature space. The SAE is optionally fine-tuned on value-task residual activations with reconstruction loss, then each value's value vector is computed from positive-minus-negative last-token SAE features. Our implementation uses pre-TopK features by default, applies a frequency threshold to reduce noisy features, and explicitly removes features shared by both polarities. At inference time, the hook encodes the residual stream, adds $\alpha$ times the value vector in the sparse SAE space, applies TopK sparsification, decodes back to the residual stream, and crucially adds back the SAE reconstruction residual error ($\Delta = a - \hat{a}$) so that non-reconstructed dense information is preserved.

\section{Geometric Fidelity vs.\ Accuracy Gain as Predictors of Cross-Value Transfer}
\label{app:predictors}

To assess which property of a steering vector better predicts its cross-value behavior, we pool all seven steering methods across both backbones (Qwen3.5-9B-Base and Llama3.1-8B), yielding $n=14$ (method, backbone) observations. For each observation, we record three quantities: geometric fidelity ($\rho_T$), target-value accuracy gain, and cross-value transfer under both TWTM and $\rho_H^{\mathrm{tr}}$. We then compute Spearman correlations between each candidate predictor and each transfer metric (Table~\ref{tab:predictors}). Geometric fidelity is the stronger predictor on both transfer metrics, and the gap is largest on $\rho_H^{\mathrm{tr}}$, which unlike TWTM, rewards only the \emph{ordering} of transfer across the Schwartz hierarchy and not steering magnitude.

\begin{table}[t]
\centering
\small
\begin{tabular}{lcc}
\toprule
\textbf{Predictor} & \textbf{TWTM} & \boldmath$\rho_H^{\mathrm{tr}}$ \\
\midrule
Geometric fidelity ($\rho_T$)  & \textbf{0.84}\rlap{$^{***}$} & \textbf{0.77}\rlap{$^{**}$} \\
Target-value accuracy gain     & 0.78\rlap{$^{***}$}          & 0.59\rlap{$^{*}$}           \\
\bottomrule
\end{tabular}
\caption{Spearman correlations between two candidate predictors and cross-value transfer, pooled across 7 steering methods $\times$ 2 backbones ($n=14$). Geometric fidelity is the stronger predictor on both transfer metrics. $^{*}p<0.05$, $^{**}p<0.01$, $^{***}p<0.001$.}
\label{tab:predictors}
\end{table}

\section{AI Assistant Acknowledgment}
During the preparation of this work, the authors utilized artificial intelligence tools to improve the grammatical clarity and flow of the manuscript. Additionally, AI-based coding assistants were used to support the implementation of the experimental pipeline and data processing scripts.

\section{Computational Resources}
All experiments were executed using a single NVIDIA B300 GPU. 

\end{document}